\documentclass[10pt,twocolumn,letterpaper]{article}

\usepackage[pagenumbers]{cvpr}

\usepackage{graphicx}
\usepackage{caption}
\usepackage{amsmath}
\usepackage{amssymb}
\usepackage{booktabs}
\usepackage{multirow}
\usepackage{makecell}
\usepackage{arydshln}
\usepackage{threeparttable}
\usepackage{cite}
\usepackage[linesnumbered,lined,vlined,ruled,commentsnumbered]{algorithm2e}
\usepackage{xcolor}

\DeclareMathOperator*{\argmin}{arg\,min}

\usepackage[pagebackref,breaklinks,colorlinks]{hyperref}

\usepackage[capitalize]{cleveref}
\crefname{section}{Sec.}{Secs.}
\Crefname{section}{Section}{Sections}
\Crefname{table}{Table}{Tables}
\crefname{table}{Tab.}{Tabs.}

\def\cvprPaperID{8568}
\def\confName{CVPR}
\def\confYear{2023}

\newcommand{\myparagraph}[1]{\noindent\textbf{#1}}

\begin{document}

\title{Class-Incremental Exemplar Compression for Class-Incremental Learning}

\author{Zilin Luo$^{1}$
\quad Yaoyao Liu$^{2}$ 
\quad Bernt Schiele$^{2}$ 
\quad Qianru Sun$^{1}$\\
\\
$^{1}$Singapore Management University \\
$^{2}$Max Planck Institute for Informatics, Saarland Informatics Campus\\
\small {
\texttt{zilin.luo.2021@phdcs.smu.edu.sg} \quad
\texttt{\{yaoyao.liu, schiele\}@mpi-inf.mpg.de}}  \quad  {\texttt{qianrusun@smu.edu.sg}}
}
\maketitle

\begin{abstract}

Exemplar-based class-incremental learning (CIL)~\cite{rebuffi2017icarl} finetunes the model with all samples of new classes but few-shot exemplars of old classes in each incremental phase, where the ``few-shot'' abides by the limited memory budget.
In this paper, we break this ``few-shot'' limit based on a simple yet surprisingly effective idea: compressing exemplars by downsampling non-discriminative pixels and saving ``many-shot'' compressed exemplars in the memory.
Without needing any manual annotation, we achieve this compression by generating $0$-$1$ masks on discriminative pixels from class activation maps (CAM)~\cite{zhou2016cam}.
We propose an adaptive mask generation model called class-incremental masking (CIM) to explicitly resolve two difficulties of using CAM: 1)~transforming the heatmaps of CAM to $0$-$1$ masks with an arbitrary threshold leads to a trade-off between the coverage on discriminative pixels and the quantity of exemplars, as the total memory is fixed; and 2)~optimal thresholds vary for different object classes, which is particularly obvious in the dynamic environment of CIL.
We optimize the CIM model alternatively with the conventional CIL model through a bilevel optimization problem~\cite{sinha2017bilevel}.
We conduct extensive experiments on high-resolution CIL benchmarks including Food-101, ImageNet-100, and ImageNet-1000, and show that using the compressed exemplars by CIM can achieve a new state-of-the-art CIL accuracy, e.g., $\textit{4.8}$ percentage points higher than FOSTER~\cite{wang2022foster} on 10-Phase ImageNet-1000.
Our code is available at \href{https://github.com/xfflzl/CIM-CIL}{https://github.com/xfflzl/CIM-CIL}.
\end{abstract}

\section{Introduction}
\label{sec_introduction}

Dynamic AI systems have a continual learning nature to learn new class data.
They are expected to adapt to new classes while maintaining the knowledge of old classes, i.e., free from forgetting problems~\cite{mcrae1993catastrophic}.
To evaluate this, the following protocol of class-incremental learning (CIL) was proposed by Rebuffi {et al}.~\cite{rebuffi2017icarl}.
The model training goes through a number of phases. Each phase has new class data added and old class data discarded, and the resultant model is evaluated on the test data of all seen classes.
A straightforward way to retain old class knowledge is keeping around a few old class exemplars in the memory and using them to re-train the model in subsequent phases.
The number of exemplars is usually limited, e.g., $5\!\sim\!20$ exemplars per class~\cite{rebuffi2017icarl, hou2019lucir, zhao2020maintaining, wu2019bic, liu2020mnemonics, douillard2020podnet, liu2021adaptive, yan2021dynamically, wang2022foster}, as the total memory in CIL strictly budgeted, e.g., $2k$ exemplars.

\begin{figure}[t]
   \center
   \includegraphics[width=0.48\textwidth]{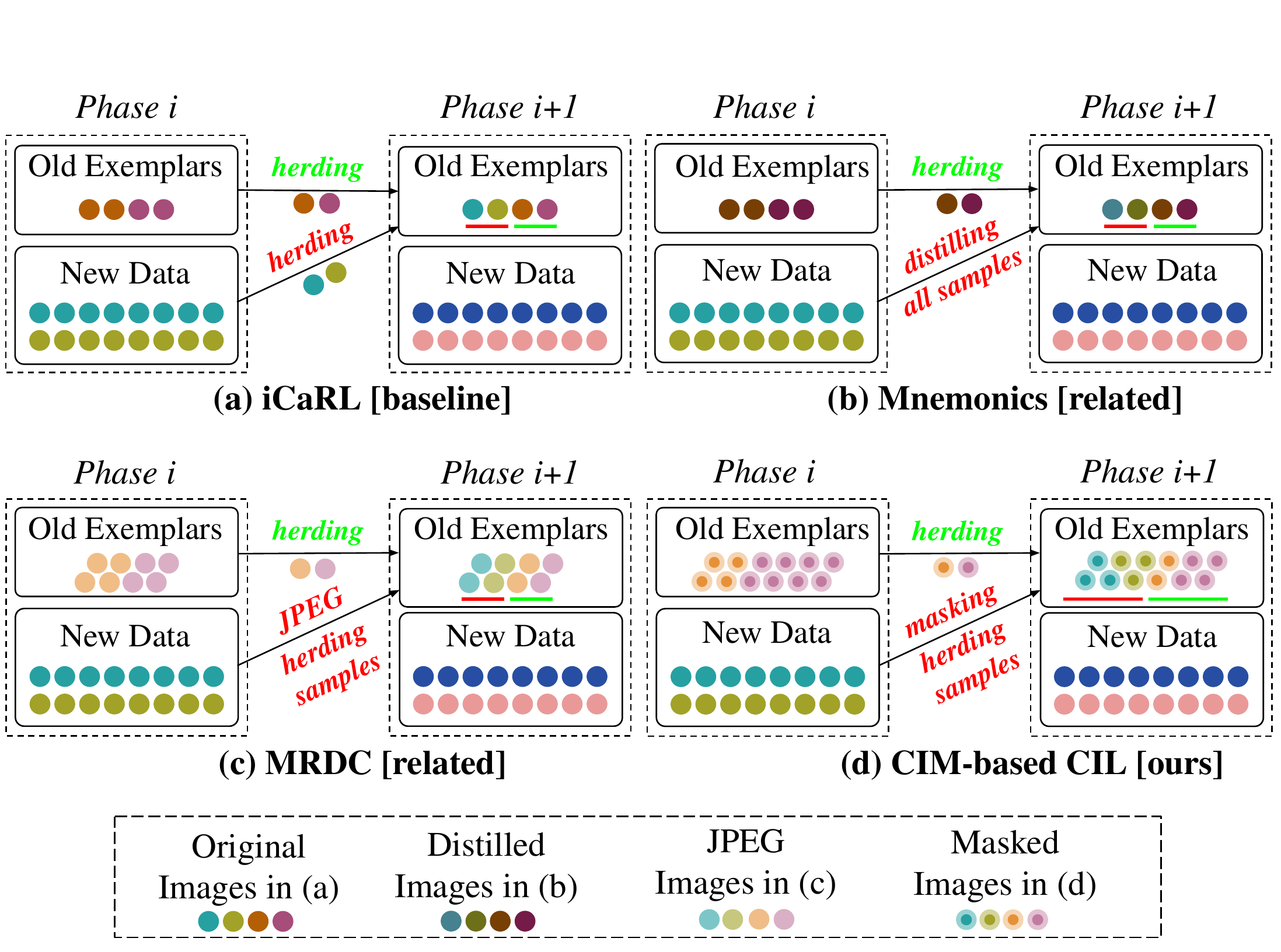}
   \caption{
   The phase-wise training data in different methods. (a) iCaRL~\cite{rebuffi2017icarl} is the baseline method using full new class data and few-shot old class exemplars.
   (b) Mnemonics~\cite{liu2020mnemonics} distills all training samples into few-shot exemplars without increasing their quantity. (c) MRDC~\cite{wang2022memory} compresses each exemplar uniformly into a low-resolution image using JPEG~\cite{wallace1991jpeg}.
   (d) Our approach based on the proposed class-incremental masking (CIM) downsamples only non-discriminative pixels in the image.
   The legend shows the symbols of special images generated by the methods.
   }
   \label{fig_teaser}
   \vspace{-0.6cm}
\end{figure}

This leads to a serious data imbalance between old and new classes, e.g., $20$ per old class \emph{vs.} $1.3k$ per new class (on ImageNet-1000~\cite{deng2009imagenet}), as illustrated in Figure~\ref{fig_teaser}\textcolor[rgb]{1,0,0}{a}.
The training is thus always dominated by new classes, and forgetting problems occur for old classes.
Liu et al.~\cite{liu2020mnemonics} tried to mitigate this problem by parameterizing and distilling the exemplars, without increasing the number of them (Figure~\ref{fig_teaser}\textcolor[rgb]{1,0,0}{b}).
Wang et al.~\cite{wang2022memory} traded off between the quality and quantity of exemplars by uniformly compressing exemplar images with JPEG~\cite{wallace1991jpeg} (Figure~\ref{fig_teaser}\textcolor[rgb]{1,0,0}{c}).
As shown in Figure~\ref{fig_teaser}\textcolor[rgb]{1,0,0}{d}, our approach is also based on image compression. The idea is to downsample only non-discriminative pixels (e.g., background) and keep discriminative pixels (i.e., representative cues of foreground objects) as the original.
In this way, \textbf{we do not sacrifice the discriminativeness} of exemplars when increasing their quantity.
In particular, \textbf{we aim for adaptive compression} in dynamic environments of CIL, where the intuition is later phases need to be more conservative (i.e., less downsampling) as the model needs more visual cues to classify the increased number of classes.

To achieve selective and adaptive compression, we need the location labels of discriminative pixels.
Without extra labeling, we automatically generate the labels by utilizing the model's own ``attention'' on discriminative features, i.e., class activation maps (CAM)~\cite{zhou2016cam}. We take this method as a feasible baseline, and based on it, we propose an adaptive version called class-incremental masking~(CIM).
Specifically, for each input image (with its class label), we use its feature maps and classifier weights (corresponding to its class label) to compute a CAM by channel-wise multiplication, aggregation, and normalization. Then, we apply hard thresholding to generate a $0$-$1$ mask.\footnote{Note that we do not use mask labels to do image compression because storing them is expensive. Instead, we expand the mask to a bounding box, as elaborated in Section~\ref{sec_methodology}.}
We notice that when generating the masks in the dynamic environments of CIL, the optimal hyperparameters (such as the value of hard threshold and the choice of activation functions) vary for different classes as well as in different incremental phases.
Our adaptive version CIM tackles this by parameterizing a mask generation model and optimizing it in an end-to-end manner across all incremental phases. 
In each phase, the learned CIM model adaptively generates class- and phase-specific masks. 
We find that the compressed exemplars based on these masks have stronger representativeness, compared to using the conventional CAM.

Technically, we have two models to optimize, i.e., the CIL model and the CIM model.\footnote{Note that the CIM model is actually a plug-in branch in the CIL model, which is detailed in Section~\ref{subsection: class_incremental_masking}.}
These two cannot be optimized separately as they are dependent on computation: 1) the CIM model compresses exemplars to input into the CIL model; 2) the two models share network parameters.
We exploit a global bilevel optimization problem (BOP)~\cite{sinha2017bilevel,chen2022gradient} 
to alternate their training processes at two levels. 
This BOP goes through all incremental training phases.
In particular, for each phase, we perform a local BOP with two steps to tune the parameters of the CIM model: 1)~a temporary model is trained with the compressed exemplars as input; and 2)~a validation loss on the uncompressed new data is computed and the gradients are back-propagated to optimize the parameters of CIM.
To evaluate CIM, we conduct extensive experiments by plugging it in recent CIL methods,\footnote{Using ``plug-in'' for evaluation is due to the fact that many baseline methods were originally evaluated in different CIL settings.}
LUCIR~\cite{hou2019lucir}, DER~\cite{yan2021dynamically}, and FOSTER~\cite{wang2022foster},
on three high-resolution benchmarks, Food-101~\cite{bossard14food101}, ImageNet-100~\cite{hou2019lucir}, and ImageNet-1000~\cite{deng2009imagenet}. 
We find that using the compressed exemplars by CIM brings consistent and significant improvements, e.g., $4.2\%$ and $4.8\%$ higher than the SOTA method FOSTER~\cite{wang2022foster}, respectively, in the $5$-phase and $10$-phase settings of ImageNet-1000, with a total memory budget for $5k$ exemplars. 

\section{Related Work}
\label{sec_related_work}

\myparagraph{Class-Incremental Learning (CIL).} 
There are three main lines of work to address the catastrophic forgetting problem~\cite{mccloskey1989catastrophic,mcrae1993catastrophic} in CIL. 
\textit{\textbf{Regularization-based}} methods apply discrepancy (between old and new models) penalization terms in their objective functions, e.g., by comparing output logits~\cite{li2017learning,rebuffi2017icarl}, intermediate features~\cite{hou2019lucir,douillard2020podnet,simon2021learning,Liu2023Online}, and prediction heatmaps~\cite{dhar2019learning}. \textit{\textbf{Parameter-isolation-based}} methods increase the model parameters in each new incremental phase, to prevent knowledge forgetting caused by parameter
overwritten.
Some of them~\cite{huang2019neural,rusu2016progressive,xu2018reinforced,yan2021dynamically,wang2022foster} proposed to progressively expand the size of the neural network to learn new coming data. Others~\cite{kirkpatrick2017overcoming,zenke2017continual,abati2020conditional,liu2021adaptive} froze a part of network parameters (to maintain the old class knowledge) to alleviate the problem of knowledge overwriting. 
\textit{\textbf{Replay-based}} methods assume there is a clear memory budget allowing a handful of old-class exemplars in the memory. Exemplars can be used to re-train the model in each new phase~\cite{rebuffi2017icarl,hou2019lucir,douillard2020podnet,liu2020mnemonics,wu2019bic,Liu2021RMM,wang2022memory}. This re-training usually contains two steps: one step of training the model on all new class data and old class exemplars, and one step of finetuning the model with a balanced subset (i.e., using an equal number of samples per class)~\cite{liu2021adaptive, Liu2021RMM, hou2019lucir, douillard2020podnet, yan2021dynamically}.

The replay-based methods focusing on \textbf{\emph{memory optimization}}~\cite{liu2020mnemonics,wang2022memory} are closely related to our work. \cite{liu2020mnemonics} proposed a bilevel optimization framework to distill the current new class data into exemplars before discarding them. It aims to improve the quality of exemplars without increasing the quantity. 
Another work~\cite{wang2022memory} aimed to trade-off between the quality and quantity of exemplars by image compression using the JPEG algorithm, i.e., each exemplar is uniformly downsampled.  
Ours differs from these two works in three aspects. 1)~Our CIM based image compression automatically segments the discriminative pixels in the exemplar and downsamples the non-discriminative pixels only.
It barely weakens the representativeness of the exemplar.
In contrast, the parameterization of image pixels in \cite{liu2020mnemonics} hampers the model from capturing high-frequency (discriminative) features from the image, especially the high-resolution (e.g., $224\!\times\!224$) image~\cite{cazenavette2022dataset}. 
2)~Our approach increases the diversity (i.e., quantity) of old class exemplars by reducing the memory consumption for each exemplar.
In contrast, \cite{liu2020mnemonics} keeps a fixed number of exemplars in the memory. 
3)~Our approach has an adaptive image compression strategy that fits well in the dynamic environments of CIL.
In contrast, \cite{wang2022memory} uses uniform image compression (or uniformly increasing quality parameters from $100$ to $1$) without considering 
the properties of specific classes in each learning phase.

\myparagraph{Class Activation Map (CAM)}~\cite{zhou2016cam} is a simple yet effective weakly-supervised object localization method. Its model is trained with only the image-level label and can generate pixel-level masks on foreground objects. Specifically, the masks are the results of hard-thresholding the heatmaps produced by feature maps and classifier weights. Advanced CAM variants include Grad-CAM~\cite{selvaraju2017grad}, ReCAM~\cite{chen2022recam}, AdvCAM~\cite{lee2021anti}, etc. 
Our CIM is based on the vanilla CAM because it is computationally simple and efficient.

\myparagraph{Bilevel Optimization Problems (BOP)}~\cite{sinha2017bilevel,chen2022gradient} aims to solve a nested optimization problem, where the outer-level optimization is subjected to the result of the inner-level optimization. It has shown effectiveness in a wide range of machine learning areas, such as hyperparameter selection~\cite{maclaurin2015gradient} and meta-learning~\cite{finn2017maml}. 
For tackling CIL tasks, \cite{liu2020mnemonics}~leverages BOP to alternatively optimize the parameters of the CIL model and the parameterized exemplars. \cite{liu2021adaptive} applies BOP to learn the aggregation
weights of the plastic and elastic branches in the CIL model. 
In our work, we use BOP to solve the optimization of the CIL model and the parameterized Class-Incremental Masking (CIM) model, where CIM is a plugin branch (in the CIL model), using few extra parameters. The process of BOP is quick and efficient.

\section{Preliminary}
\label{sec_an_overview_of_cil}

The following is the training pipeline of standard CIL with few-shot exemplars.
Assume there are $N$ learning phases. 
In the $1$-st phase, we load data $\mathcal{D}_{1}$ containing all training samples of $c_1$ classes, and use $\mathcal{D}_{1}$ to train the initial classification model $(\theta_1, \omega_1)$, where $\theta_1$ and $\omega_1$ 
 denote the parameters of the feature extractor and classifier, respectively. 
When the training is done, we evaluate the model performance on the test samples of $c_1$ classes. Before the $2$-nd phase, we discard most of the training samples due to the strict memory budget of CIL.
In other words, we preserve only a handful of training samples $\mathcal{E}_1$ (i.e., exemplars) in the memory, selected from $\mathcal{D}_1$. A common method for selecting exemplars is called feature herding~\cite{rebuffi2017icarl} and has been used in many related works~\cite{liu2020mnemonics,yan2021dynamically,wang2022memory,wang2022foster}. We adopt it, too, in this work.
In the $i$-th phase ($i\geq2$), we load all exemplars ${\mathcal E}_{1:i-1}=\mathcal{E}_1\cup \dots\cup {\mathcal E}_{i-1}$ from the memory and initialize the current model $(\theta_{i},\omega_{i})$ by the previous model $(\theta_{i-1},\omega_{i-1})$. 
We use $\mathcal{E}_{1:i-1}$ and the new coming data $\mathcal D_i$ (containing $c_i$ new classes) to train $(\theta_{i},\omega_{i})$.
Then, we evaluate the current model using a test set of all $\sum_{j=1}^{i}c_j$ classes seen so far. 
After that, we discard most of the training samples in $\mathcal D_i$, and leave few-shot exemplars ${\mathcal E}_{i}$ in the memory. 
It is clear that this discarding causes a strong data imbalance between old and new coming classes in the subsequent phase. In the following, we introduce our solution to this problem.

\section{Methodology}
\label{sec_methodology}

\begin{figure*}[t]
\centering
\vspace{-0.2cm}
\includegraphics[width=\textwidth]{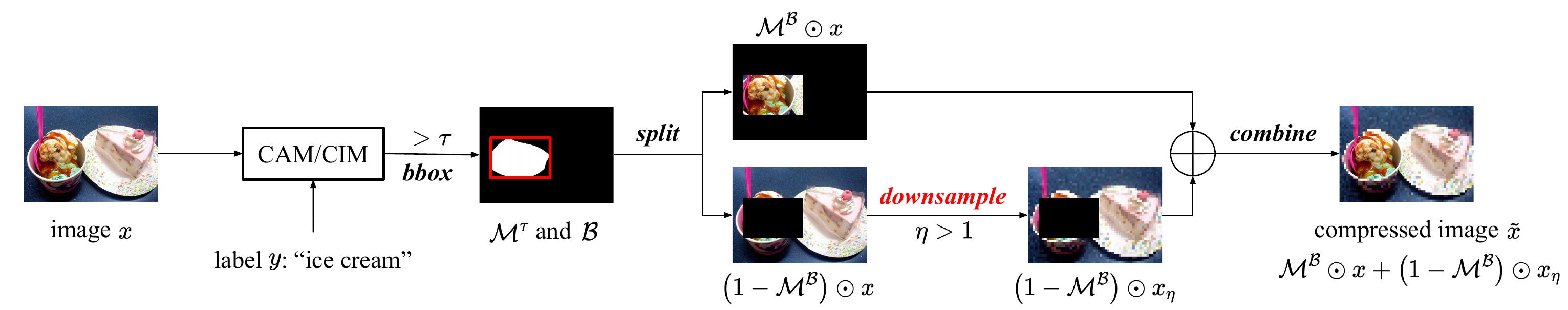}
\caption{The proposed compression pipeline.
Given an image, we extract its CAM-based (baseline) or CIM-based (ours) mask, threshold it to be a 0-1 mask with a fixed threshold $\tau$, and generate a tight bounding box (bbox) that covers all masked pixels. Then, we downsample the pixels outside the bbox and sum the downsampled image up to the masked image to generate the final compressed image.}
\vspace{-0.4cm}
\label{figure: framework}
\end{figure*}

As illustrated in Figure~\ref{fig_teaser}\textcolor[rgb]{1,0,0}{d}, we alleviate the data imbalance problem by saving a larger number of compressed exemplars for old classes, where
we leverage pixel-selective compression, i.e., downsample only non-discriminative pixels.
To achieve this, \textbf{the ideal case} is that we have the pixel-level localization of foreground objects.
However, \textbf{the realistic case} is that such localization labels are expensive, most CIL benchmarks do not have the labels, and it is not fair to compare with other CIL methods if using the labels.
Without extra labeling, we introduce a CAM-based mask generation method, and based on it, we provide a baseline solution to pixel-selective compression in Section~\ref{subsection: cam_based_compression_pipeline}.
\textbf{The problem} of mask generation in CIL is that the optimal generation hyperparameters such as hard thresholds are changing in the dynamic environment (with the increasing number of classes and phases).
It is thus desirable to have an adaptive mask generation process.
To this end, we propose class-incremental masking (CIM)---a learnable mask generation model, in Section~\ref{subsection: class_incremental_masking}.

\subsection{CAM-based Compression Pipeline}
\label{subsection: cam_based_compression_pipeline}

Generating pixel-level labels for large-scale datasets, e.g., ImageNet~\cite{deng2009imagenet}, is non-trivial.
Using class activation maps (CAM) is a na\"ive solution with little computation costs. Its key idea is to make use of the activation of the classification model itself: on the feature maps, activated pixels are more discriminative than non-activated ones for recognizing the object, where ``activated'' means \emph{of high activation values} and \emph{with strong correlation with the classification weights of the object}.
After localizing activated pixels by CAM, we can generate a $0$-$1$ mask on them, e.g., by hard thresholding their normalized values, and then upsample the mask to the size of the input image~\cite{zhou2016cam}.

\myparagraph{From CAM to $0$-$1$ Mask.}
We extract CAM in the following steps. Given an image $x$ from $\mathcal{D}_i$ and its ground truth class label $y$, let $F(x;\theta_i)$ represent the feature block output by the feature extractor $\theta_i$, and $\omega_{i,y}$ for the classification weights of class $y$ in the classifier $\omega_i$.
The CAM of $x$ is:
\begin{equation}
\label{eq_cam}
\mathcal{M}^\textrm{CAM}=\frac{A-\min{(A)}}{\max{(A)} - \min{(A)}}, \ A={\omega_{i,y}^{\top}F(x;\theta_i)},
\end{equation}
where $\min(\cdot)$ and $\max(\cdot)$ operations are used for normalization.
Then, we upsample $\mathcal{M}^\textrm{CAM}$ to the size of image $x$ and use the same notation.
Each value in $\mathcal{M}^\textrm{CAM}$
denotes the activation strength of the model at a specific pixel location.
Following the way of generating $0$-$1$ masks in weakly-supervised semantic segmentation works~\cite{dong_2020_conta, chen2022recam, lee2021anti}, 
we apply a hard threshold $\tau$ (between $0$ and $1$) over all values of $\mathcal{M}^\textrm{CAM}$, and get the $0$-$1$ mask $\mathcal{M}^\tau$:
$\mathcal{M}^\tau\!=\!\mathbb{I}(\mathcal{M}^\textrm{CAM}\!>\!\tau)$, where $\mathbb{I}(\cdot)$ is the indicator function. 
In $\mathcal{M}^\tau$, $1$s indicate the locations of discriminative pixels, e.g., foreground pixels, based on which the model makes the prediction. 
While $0$s indicate mostly background pixels or non-discrimination pixels that can be downsampled as they make little contribution to the prediction.

After generating $0$-$1$ masks, it is ideal to keep them in the memory as meta information of compression. However, this is not efficient or feasible in CIL. There are two reasons.
1) The space for saving image-size masks is non-negligible.
Each mask pixel is a one-bit boolean value, and one mask takes around $\frac{1}{3\times8}=\frac{1}{24}$ of the memory of one RGB image.
2) The mask involves activated regions with irregular shapes. It is thus non-trivial to perform any standard downsampling algorithm~\cite{parker1983comparison} on the remaining regions.

\noindent
\textbf{From $0$-$1$ Masks to Bounding Boxes (BBox).}
A simple workaround is to generate a tight bounding box (bbox) to cover the positions of $1$s in $\mathcal{M}^\tau$, and use the bbox for compression.
Specifically, given $\mathcal{M}^\tau$, we obtain the coordinate representation of the bounding box as:
\begin{equation}
\label{eq_bbox}
\mathcal{B}=[\min h,\min w;\max h,\max w]_{(h,w):\mathcal{M}^\tau(h,w)=1},
\end{equation}
where $h$ and $w$ denote the vertical and horizontal coordinates of the $1$ on $\mathcal{M}^\tau$, respectively.
We highlight that $\mathcal{B}$ consists of four integers only and takes negligible memory overhead compared to $\mathcal{M}^\tau$. In addition, we ``reshape'' the irregular shape (of the activated region) in $\mathcal{M}^\tau$ into a rectangular $\mathcal{B}$, so our downsampling operation on the pixels outside the rectangular becomes easy.

\myparagraph{Compression with BBox.}
Given the image and its bbox on foreground, the compression is implemented by downsampling pixels outside the bbox.
Specifically, as illustrated in Figure~\ref{figure: framework}, we compress the image $x$ to $\tilde{x}$ as follows,
\begin{equation}
\label{eq_compressed_image}
\tilde{x}=\mathcal{M}^\mathcal{B} \odot x + (1 - \mathcal{M}^\mathcal{B}) \odot x_\eta,
\end{equation}
where $\mathcal{M}^\mathcal{B}$ is the binary mask according to $\mathcal{B}$, i.e., the values of $\mathcal{M}^\mathcal{B}$ are $1$ inside $\mathcal{B}$, and $0$ otherwise. 
$x_\eta$ is the fully downsampled version of $x$ with a downsampling ratio $\eta$ ($\eta>1$), $\odot$ denotes the element-wise product,
and $+$ denotes the element-wise addition. Both $\odot$ and $+$ are applied independently on each RGB channel.

The memory allocated for the compressed image $\tilde{x}$ is as follows,
\begin{equation}
\begin{aligned}
\label{eq_consumed_memory}
m_{\tilde{x}}&=\frac{H_{\mathcal{B}}W_{\mathcal{B}}}{HW} + \frac{1}{\eta}\Big(1 - \frac{H_{\mathcal{B}}W_{\mathcal{B}}}{HW}\Big)\\&=1-\Big(1-\frac{1}{\eta}\Big)\cdot\Big(1-\frac{H_\mathcal{B}W_\mathcal{B}}{HW}\Big),
\end{aligned}
\end{equation}
where $H_{\mathcal{B}}$ and $W_{\mathcal{B}}$ are the height and width of $\mathcal{B}$, respectively. $H$ and $W$ are the height and width of the original image $x$, respectively.
$m_{\tilde{x}}$ is always smaller than $1$ where $1$ denotes the memory unit of saving one original image $x$. Therefore, we can save a larger number of compressed exemplars within the same memory budget. We denote the set of compressed exemplars in the $i$-th phase as $\tilde{{\mathcal E}}_i$.

\myparagraph{Compression Artifacts.} 
The above compression introduces artifacts to the compressed images,
i.e., there is a resolution mutation around bounding box edges.
From the perspective of spectrum analysis~\cite{jain1989fundamentals,castleman1996digital}, such mutation carries noisy and high-frequency components and impairs the model training in subsequent phases. 
We mitigate the effect of these artifacts by implementing the following data augmentation: in each training epoch, we transform a random subset of $\mathcal{D}_i$ into compressed images with CAM-based bounding boxes using the same downsampling ratio.
Using this augmentation enables the model to ``simulate'' the training with compressed images and learn to be invariant to compression artifacts.

\subsection{Class-Incremental Masking (CIM)}
\label{subsection: class_incremental_masking}

Ideally, the mask generation process needs to adjust at different phases in CIL environments. 
The process involves two hyperparameters: the masking threshold and the choice of network activation functions.
First, for the threshold, searching for its optimum (for all classes) is not trivial.
Grid search is intuitive, but it is computationally expensive when the number of classes increases in CIL.
Second, for the activation function, the standard network of CIL methods uses ReLU~\cite{nair2010rectified,rebuffi2017icarl} and does not optimize it.
We solve this issue by applying a learnable activation function in addition to the existing ReLU function in the CIL model.
Physically, we have one neural network, while logically, we have two models to learn (in each incremental phase): the conventional CIL model with ReLU activations, and the adaptive mask generation model with learnable activations.
We thus call our method class-incremental masking (CIM) based CIL. In the following, we elaborate on the network design and optimization pipeline.

\myparagraph{Network Design.}

\begin{figure}
   \vspace{-0.5cm}
   \center
   \includegraphics[width=0.48\textwidth]{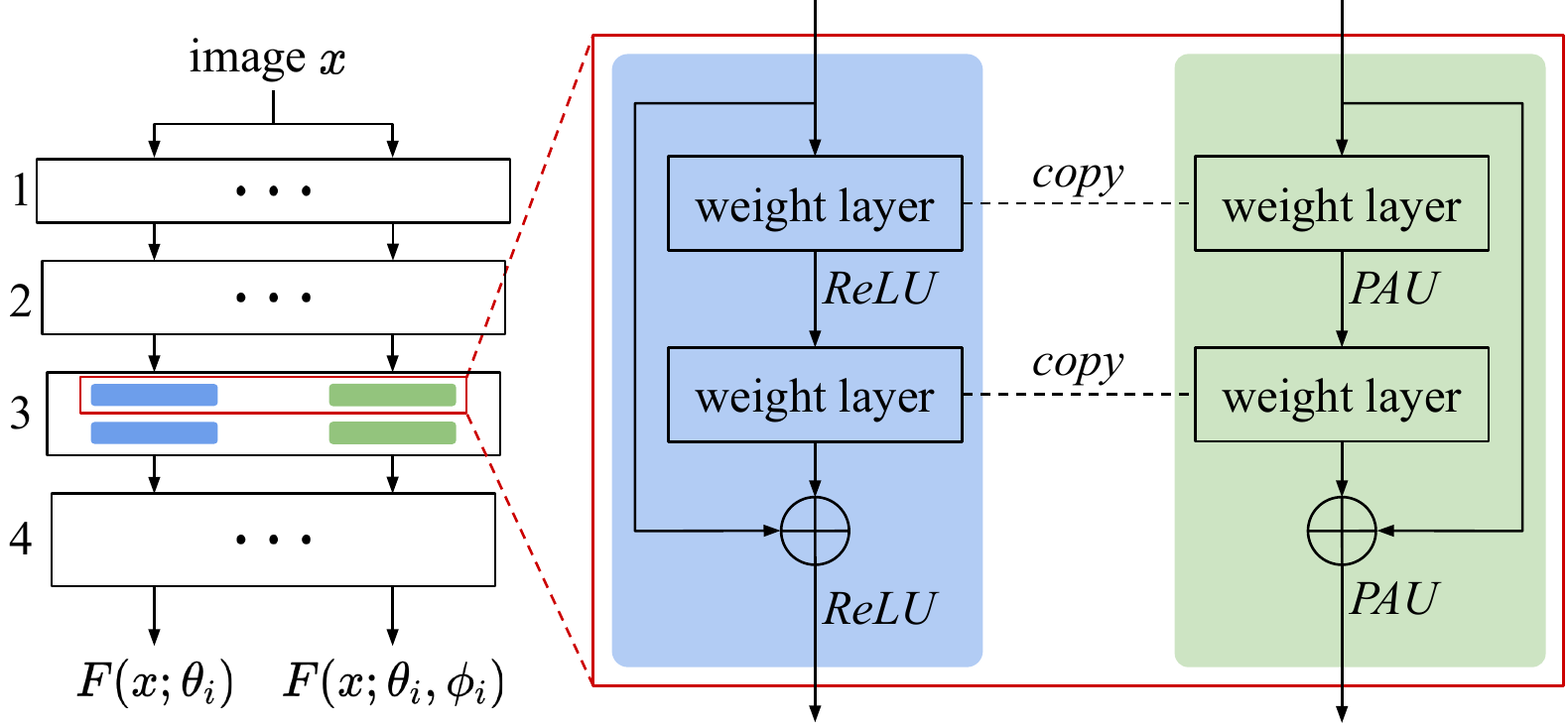}
   \vspace{-0.3cm}
   \caption{
   Our CIM installs an ``extension'' on the backbone (e.g., ResNet-18~\cite{he2016deep,rebuffi2017icarl} with four blocks) by adding a learnable activation function (e.g., PAU~\cite{molina2019pade}) at the position of the original activation function (i.e., ReLU~\cite{nair2010rectified}). Such ``extension'' results in a new network branch (in \textcolor[rgb]{0.576, 0.769, 0.490}{green}), whose weight layer parameters are directly copied from the original branch (in \textcolor[rgb]{0.427, 0.620, 0.922}{blue}).
   }
   \label{figure: revised_activation_layer}
   \vspace{-0.4cm}
\end{figure}

Figure~\ref{figure: revised_activation_layer} demonstrates an example network architecture in our CIM-based CIL.
The proposed CIM extends the network backbone by \emph{logically} adding a network branch, where only the activation functions are learnable (e.g., Pad{\'e} Activation Units (PAU\footnote{PAU uses a rational function of given degrees $m$ and $n$, i.e., $(a_0+a_1x+\cdots+a_mx^m)/(1+b_1x+\cdots+b_nx^n)$, and can be parameterized by $a_0,\cdots,a_m$ and $b_1,\cdots,b_n$.},~\cite{molina2019pade})) and the parameters of weight layers are copied from the original branch.

This design is motivated by the works of He et al.~\cite{he2015prelu} and Bochkovskiy et al.~\cite{bochkovskiy2020yolov4}, which indicate that layers with learnable activation functions can flexibly process object (localization) information at different network blocks. The difference is that we apply this flexibility to achieve adaptive mask generation for different CIL phases. 

We denote the CIM parameters (i.e., the parameters in learnable activation functions) in $i$-th phase as $\phi_i$.
We optimize the CIL model $(\theta_i, \omega_i)$ and the CIM model $\phi_i$ via a global BOP, as elaborated below.

\begin{algorithm}
\caption{CIM-based CIL ($i$-th phase, $i\!\geq\!1$)} 
\label{algorithm: optimization flow}
\SetAlgoLined
\SetKwInput{KwData}{Input}
\SetKwInput{KwResult}{Output}
\KwData{New data $\mathcal{D}_i$; old compressed exemplars $\tilde{{\mathcal E}}_{0:i-1}$ ($\tilde{{\mathcal E}}_{0}=\varnothing$); last-phase CIL model $(\theta_{i-1},\omega_{i-1})$ and CIM model $\phi_{i-1}$ ($\theta_0$, $\omega_0$ are random parameters, $\phi_0$ is set to ReLU).}
\KwResult{New compressed exemplars $\tilde{{\mathcal E}}_i$; updated CIL model $(\theta_{i},\omega_{i})$ and CIM model $\phi_i$.}
Initialize $(\theta_{i},\omega_{i})$ with $(\theta_{i-1},\omega_{i-1})$\;
Initialize $\phi_i$ with $\phi_{i-1}$\;
\For{\textrm{epochs}}{
    Train $(\theta_{i},\omega_{i})$ using $\tilde{{\mathcal E}}_{0:i-1}\cup\mathcal{D}_i$ by Eq.~\ref{eq_cil_training}\;
    Compress $\mathcal{D}_i$ into $\tilde{\mathcal{D}}^{\phi}_i$ using $\phi_i$\; 
    Temporarily update $\theta_{i}$ to $\theta_{i}^+$ using $\tilde{{\mathcal E}}_{0:i-1}\cup\tilde{\mathcal{D}}^{\phi}_i$ by Eq.~\ref{eq_inner_level_update}\;
    Learn $\phi_i$ using $\mathcal{D}_i$ by Eq.~\ref{eq_outer_update_new}\;
}
Compress $\mathcal{D}_i$ into $\tilde{\mathcal{D}}_i$ using the learned $\phi_i$ by Eq.~\ref{eq_compressed_image}\;
Select exemplars $\tilde{\mathcal{E}}_i$ from $\tilde{\mathcal{D}}_i$ by, e.g., herding~\cite{rebuffi2017icarl}.
\end{algorithm}

\noindent
\textbf{Optimization Pipeline.}
We demonstrate the overall optimization flow in Algorithm~\ref{algorithm: optimization flow}, which consists of two levels of optimization: task-level and mask-level---the former one for CIL and the latter one for CIM. Note that to maintain an unified notation, we further define $\tilde{{\mathcal E}}_{0}=\varnothing$.

\myparagraph{1) Task-level Optimization.}
This level aims to optimize the CIL model $(\theta_i,\omega_i)$ to address the CIL task at hand. It can be written as: 
\begin{equation}
\begin{aligned}
\label{eq_cil_training}
(\theta_{i},\omega_{i})\leftarrow(\theta_{i},\omega_{i})-\lambda\nabla_{(\theta,\omega)}\mathcal{L}_{\textrm{CIL}}(\tilde{\mathcal{E}}_{0:i-1}\!\cup\!\mathcal{D}_i;\theta_{i},\omega_{i}),
\end{aligned}
\end{equation}
where $\lambda$ is the learning rate. We follow the implementation of CIL training loss $\mathcal{L}_{\textrm{CIL}}$ in baseline methods~\cite{li2017learning,zhou2021co,yan2021dynamically,wang2022foster}. This means that we use different training losses when plugging CIM into different baseline methods.

\myparagraph{2) Mask-level Optimization.}
This level aims to optimize the CIM model $\phi_i$ to produce adaptive compression masks. It is formulated as a local BOP:
\begin{subequations}
\begin{align}
&\min_{\phi_i}\left[{\mathcal{L}_{\textrm{val}}(\mathcal{D}_i;\theta^*_i,\omega_i)+\mu\mathcal{R}(\phi_i)}\right]\label{eq_bop_formulation_outer},\\
&~\mathrm{s.t.}~\theta_i^*=\argmin_{\theta_i}{\mathcal{L}_\textrm{train}(\tilde{\mathcal{E}}_{0:i-1}\cup\tilde{\mathcal{D}}_i(\phi_i);\theta_i,\omega_i)}. \label{eq_bop_formulation_inner}
\end{align}
\end{subequations}
Eq.~\ref{eq_bop_formulation_inner} denotes an inner-level optimization. It trains $\theta_i$ with the data $\tilde{\mathcal{D}}_i(\phi_i)$ compressed by using $\phi_i$, and converges as $\theta_i^*$.
Eq.~\ref{eq_bop_formulation_outer} denotes an outer-level optimization. It is based on the validation loss derived by $\theta_i^*$ on the original data $\mathcal{D}_i$.
$\mathcal{R}(\phi_i)$ is a constraint representing the memory limitation and $\mu$ is its weight.
In the following, we elaborate on the implementation details for the two levels.

In the inner-level optimization, we train a temporary CIL model with compressed data. 
Specifically, we first compress new-class data $\mathcal{D}_i$ into $\tilde{\mathcal{D}}_i(\phi_i)$ using the generated masks by the CIM model $\phi_i$.
Then, we implement the inner-level optimization as a one-step gradient descent (using the CIL training loss) as:
\begin{equation}
\label{eq_inner_level_update}
\theta^+_i\leftarrow\theta_i-\beta_1\nabla_{\theta}\mathcal{L}_\textrm{CIL}(\tilde{\mathcal{E}}_{0:i-1}\cup\tilde{\mathcal{D}}_i(\phi_i);\theta_i,\omega_i),
\end{equation}
where $\beta_1$ is the learning rate for $\theta_i$. 

The aim of the outer-level optimization is to optimize $\phi_i$ such that the temporary model $(\theta^+_i,\omega_i)$ (trained with compressed data) has a low validation loss on the original data $\mathcal{D}_i$.
To achieve this, we back-propagate the loss on the original data~$\mathcal{D}_i$
to update $\phi_i$ as:
\begin{equation}
\label{eq_outer_level_update}
\phi_i\leftarrow\phi_i-\beta_2\nabla_{\phi}\left[\mathcal{L}_{\textrm{CE}}(\mathcal{D}_i;\theta^+_i,\omega_i)+\mu\mathcal{R}(\phi_i)\right],
\end{equation}
where $\mathcal{L}_{\textrm{CE}}$ denotes softmax cross-entropy loss and $\beta_2$ is the learning rate for $\phi_i$. 
This trains $\phi_i$ to capture the most discriminative features of new-class images.
The constraint $\mathcal{R}(\phi_i)$ is 
implemented as a $\ell_2$-regularization term on the generated mask by $\phi_i$. The motivation for the regularization term is to make the mask coverage smaller, thus the compressed images take less memory.

We empirically observe that by the above optimization flow, the output activation maps
by $\phi_i$ are easy to collapse, i.e., different images have the same map.
To solve this issue, we add a cross-entropy loss term about $\phi_i$ in Eq.~\ref{eq_outer_level_update} to regularize it to produce image-specific activation maps:
\begin{equation}
\begin{aligned}
\label{eq_outer_update_new}
\phi_i\leftarrow\phi_i-\beta_2\nabla_{\phi}[\mathcal{L}_{\textrm{CE}}(&\mathcal{D}_i;\theta^+_i,\omega_i)+\mu\mathcal{R}(\phi_i)\\+\mu^\prime&\mathcal{L}_{\textrm{CE}}(\tilde{\mathcal{E}}_{0:i-1}\cup\mathcal{D}_i;\theta_i,\phi_i,\omega_i)],
\end{aligned}
\end{equation}
where $\mu^\prime$ is the weight. 

\myparagraph{Limitations.} Our CIM learns to generate adaptive masks for exemplar compression in CIL. It has three limitations that are left as future work.
1) It is not able to adjust any previous-phase exemplars, as the validation data (the original data of these exemplars) are not accessible anymore.
2) It introduces hundreds of activation parameters to the CIL model, although this is not a significant overhead compared to model parameters.
Please check detailed overhead analyses in the supplementary materials.
3) Image compression is not that meaningful for low-resolution datasets (e.g., $32\!\times\!32$ CIFAR-100)
It is because the memory taken by the compression parameters (e.g., the parameters of CIM) and the RGB pixels of a low-resolution image are comparable. 
Using the memory to save more images is more meaningful.
\section{Experiments}
\label{sec_experiments}

We incorporate CIM into two baseline CIL methods (i.e., DER~\cite{yan2021dynamically} and FOSTER~\cite{wang2022foster}) and boost their model performances consistently on three datasets.
Below, we introduce datasets and experiment settings (Section~\ref{subsection: experimental_settings}), followed by results and analyses (Section~\ref{subsection: results_and_analyses})

\subsection{Experimental Settings}
\label{subsection: experimental_settings}

\myparagraph{Datasets.}
We conduct experiments on three standard CIL benchmarks with high-resolution images.
1)~\textbf{Food-101}~\cite{bossard14food101} consists of $101$ food categories with $750$ training and $250$ test samples per category.
All images have a maximum side length of $512$ pixels.
2)~\textbf{ImageNet-1000}~\cite{deng2009imagenet} is a large-scale dataset with $1,\!000$ classes and each class has around $1,\!300$ training and $50$ test samples.
3)~\textbf{ImageNet-100} is a 100-class subset randomly sampled from ImageNet-1000 with a fixed NumPy~\cite{harris2020array} random seed ($1993$), following~\cite{hou2019lucir}.
We provide other details of these datasets, e.g., image sizes and pre-processing methods, in the supplementary materials.

\myparagraph{Protocols.}
We use two protocols: \textit{learning from scratch} (LFS) and \textit{learning from half} (LFH), following recent CIL works~\cite{yan2021dynamically,wang2022foster}.
In LFS, the model observes the same number of classes in all $N$ phases, where $N$ is optionally $5$, $10$, and $20$. 
In LFH, the model is trained on half of the classes (e.g., $500$ classes for ImageNet-1000) in the $1$-th phase. Then, it learns the remaining classes evenly in the subsequent $N$ phases, where $N$ can be $5$, $10$, and $25$.
In both protocols, after the training of each phase we evaluate the resultant model on the test data of all seen classes.
Our final report includes the average accuracy over all phases and the last-phase accuracy which indicates the degree of model forgetting.
We run each experiment three times and report the average results.

\myparagraph{Memory Budget.}
There are two memory budget\footnote{Please note that we measure the memory by the number of original images. Each compressed exemplar in CIM takes less memory than the original image (Eq.~\ref{eq_consumed_memory}), resulting in more exemplars in the same memory.} settings. 1)~In the ``fixed'' setting, we remove some old-class exemplars when new exemplars from the current phase are added in the memory to maintain the ``fixed memory budget''. In this setting, we set the total memory to be $2,\!020$ samples for Food-101 and $2,\!000$ samples for ImageNet-100. For ImageNet-1000, we have two options---$5,\!000$ samples and $20,\!000$ samples. 2)~In the ``growing'' setting, a constant memory budget is allocated for each class across all phases and hence extra memory is appended when new classes come. In this setting, we set the budget to be $20$ samples per class for all datasets. Following~\cite{yan2021dynamically,wang2022foster}, we apply the ``fixed'' setting in LFS experiments and the ``growing'' setting in LFH experiments.

\begin{table*}[t]
\normalsize
\begin{center}
\renewcommand\arraystretch{1}
\setlength{\tabcolsep}{1.6mm}{
\begin{threeparttable}
\begin{tabular}{lccccccccccccccc}
\toprule
\multirow{3.5}{*}{\textbf{Method}}
& \multicolumn{7}{c}{\textit{Learning from Scratch (LFS)}} && \multicolumn{7}{c}{\textit{Learning from Half (LFH)}} \\
\cmidrule{2-8} \cmidrule{10-16}
& \multicolumn{3}{c}{\textit{Food-101}} && \multicolumn{3}{c}{\textit{ImageNet-100}} && \multicolumn{3}{c}{\textit{Food-101}} && \multicolumn{3}{c}{\textit{ImageNet-100}} \\
\cmidrule{2-4} \cmidrule{6-8} \cmidrule{10-12} \cmidrule{14-16}
& $N$=5 & 10 & 20 && 5 & 10 & 20 && 5 & 10 & 25 && 5 & 10 & 25 \\
\hline
iCaRL~\cite{rebuffi2017icarl}               & 69.66 & 62.18 & 56.70 && 73.90 & 67.06 & 62.36 && 60.13 & 53.42 & 46.87 && 62.53 & 59.88 & 52.97  \\
WA~\cite{zhao2020maintaining}               & 70.94 & 63.69 & 58.45 && 74.64 & 68.62 & 63.20 && 63.55 & 57.60 & 52.48 && 65.75 & 63.71 & 58.34  \\
PODNet~\cite{douillard2020podnet}           & 68.03 & 61.24 & 47.38 && 72.14 & 63.96 & 53.69 && 75.37 & 70.01 & 65.32 && 75.54 & 74.33 & 68.33  \\
AANets~\cite{liu2021adaptive}               & 69.46 & 61.59 & 48.83 && 72.98 & 65.77 & 55.36 && 76.07 & 71.22 & 66.93 && 76.96 & 75.58 & 71.78  \\
\cdashline{1-16}[4pt/2pt]
DER~\cite{yan2021dynamically}               & 73.88 & 70.76 & 64.39 && 78.50 & 76.12 & 73.79 && 78.13 & 73.45 &   -   && 79.08 & 77.73 &   -    \\
DER~\textit{w}/ ours                        & 75.63 & 73.09 & 69.17 && 79.63 & 77.57 & \textbf{75.36} && 79.25 & 75.76 &   -   && 80.30 & \textbf{79.05} &   -    \\
\cdashline{1-16}[4pt/2pt]
FOSTER~\cite{wang2022foster}   & 75.03 & 72.72 & 66.73 && {79.93}\tnote{\dag} & {76.55}\tnote{\dag} & 74.49 && 79.08 & 75.07 & 68.08 && {80.07}\tnote{\dag} & 77.54 & 72.40\tnote{*}  \\
FOSTER~\textit{w}/ ours                     & \textbf{76.44} & \textbf{74.85} & \textbf{70.20} && \textbf{80.58} & \textbf{77.94} & 75.23 && \textbf{79.76} & \textbf{76.86} &\textbf{70.50} && \textbf{80.93} & 78.66 & \textbf{75.74}  \\
\bottomrule
\end{tabular}
{\small
\begin{tablenotes}
\item[\dag] The paper of FOSTER~\cite{wang2022foster} did not report the numerical results for $N$=$5$/$10$ (LFS) and $N$=$5$ (LFH) on ImageNet-100. We run these experiments using the public code (released by authors) and report the reproduced results.
\item[*] Our reproduced result ($72.40$) is significantly higher than the original result ($69.34$) reported in the paper of FOSTER.
\end{tablenotes}
}
\end{threeparttable}}
\end{center}
\vspace{-5mm}
\caption{Average accuracies (\%) of two top-performing CIL methods~\cite{yan2021dynamically,wang2022foster} with and without our CIM-CIL plugged-in, and other four baselines~\cite{rebuffi2017icarl,zhao2020maintaining,douillard2020podnet,liu2021adaptive}, on two datasets (Food-101 and ImageNet-100) and using two protocols (learning from scratch (LFS) and learning from half (LFH)). Due to the space limits, we report the $95\%$ confidence intervals for these results in the supplementary materials.
}
\vspace{-0.4cm}
\label{table: sota_food101_and_imagenet100}
\end{table*}

\myparagraph{Implementation Details.}
Our implementation is based on the standard deep learning library PyTorch~\cite{paszke2019pytorch} and image processing library OpenCV~\cite{mordvintsev2014opencv}. Following~\cite{yan2021dynamically,wang2022memory,wang2022foster}, we use an $18$-layer ResNet~\cite{he2016deep} as the network backbone $\theta$ and a fully-connected layer as the classifier $\omega$ in all experiments. We use the same CIL training hyperparameters as in related works~\cite{hou2019lucir,yan2021dynamically,wang2022foster} for fair comparison: 1)~there are $200$ epochs in $1$-st phase and $170$ epochs in the subsequent phases; 2)~the learning rate $\lambda$ is initialized as $0.1$ and decreases to zero with a cosine annealing scheduler~\cite{loshchilov2016sgdr}; 3)~the SGD optimizer is deployed, with momentum factor set to $0.9$ and weight decay set to $0.0005$. For compression-related hyperparameters, we set the masking threshold $\tau$ as $0.6$ and the downsampling ratio $\eta$ as $4.0$. To build the CIM model, we apply PAUs with degrees $m=5$ and $n=4$ as learnable activation layers. For the optimization of the CIM model $\phi$ (i.e., the mask-level optimization), we initially set $\beta_1$ as $0.1$ and $\beta_2$ as $0.01$ and reduce them to zero following the scheduler of $\lambda$. $\mu$ and $\mu^\prime$ is set to $0.1$ and $0.2$, respectively. To smooth the training, we clip the gradient norm of $\phi$ to be no more than $1$. \emph{We report the result of hyperparameter sensitivity analysis in the supplementary materials.}

\subsection{Results and Analyses}
\label{subsection: results_and_analyses}

\begin{table}[t]
\normalsize
\begin{center}
\renewcommand\arraystretch{1}
\setlength{\tabcolsep}{1.0mm}{
\begin{tabular}{llccccccc}
\toprule
\multirow{2.3}{*}{\makecell{\textbf{Memory}\\\textbf{Budget}}} & \multirow{2.5}{*}{\textbf{Method}} & \multicolumn{2}{c}{$N$=5} && \multicolumn{2}{c}{$N$=10} \\
\cmidrule{3-4} \cmidrule{6-7}
&& Avg. & Last && Avg. & Last \\
\hline
\multirow{5}{*}{$M\!=\!20k$}
& iCaRL~\cite{rebuffi2017icarl}                                     & 44.36 & 27.78 && 38.40 & 22.70  \\
& WA~\cite{zhao2020maintaining}                                     & 58.37 & 50.62 && 54.10 & 45.66  \\
& DER~\cite{yan2021dynamically}                                     & 67.49 & 59.75 && 66.73 & 58.62  \\
\cdashline{2-9}[4pt/2pt]
& FOSTER~\cite{wang2022foster}                                      & 69.21 & 64.88 && 68.34 & 60.14  \\
& FOSTER~\textit{w}/ ours                                           & \textbf{69.93} & \textbf{66.05} && \textbf{69.53} & \textbf{62.07}  \\
\cdashline{1-9}[4pt/2pt]
\multirow{2}{*}{$M\!=\!5k$}
& FOSTER                                                            & 57.19 & 49.42 && 54.72 & 44.96  \\
& FOSTER~\textit{w}/ ours                                           & \textbf{61.37} & \textbf{54.46} && \textbf{59.48} & \textbf{50.83}  \\
\bottomrule
\end{tabular}}
\end{center}
\vspace{-4mm}
\caption{Average and last accuracies (\%) on ImageNet-1000 of FOSTER with and without our method plugged-in, and other three baselines~\cite{rebuffi2017icarl,zhao2020maintaining,yan2021dynamically}. We show two memory budgets, $M\!=\!20k$ (upper block) and $M\!=\!5k$ (lower block), in the LFS setting.}
\vspace{-4mm}
\label{table: sota_imagenet1000}
\end{table}
\myparagraph{Comparing with the State-of-the-art.}
In Table~\ref{table: sota_food101_and_imagenet100}, we summarize the experimental results on two datasets (Food-101 and ImageNet-100) and in two CIL protocols (LFS and LFH). From the table, we have the following observations. 1)~Our CIM-based CIL consistently improves the state-of-the-art method FOSTER~\cite{wang2022foster} with clear margins in all settings. E.g., our method surpasses it by an average of $1.4$ percentage points on ImageNet-100, and $2.0$ percentage points on Food-101. 
2) Our CIM-based CIL achieves more significant improvements when $N$ becomes larger, e.g., on ImageNet-100 (LFH), our method improves FOSTER by $0.9$ and $3.3$ percentage points when $N$=$5$ and $N$=$25$, respectively. 3) Our CIM-based CIL achieves greater improvements consistently on Food-101 (than ImageNet-100). It improves baselines by $2.1$ percentage points on Food-101, while the improvement is $1.4$ on ImageNet-100 ($N$=$10$, LFS). 
It shows that our method is particularly effective when the representative visual cues of a class are from some of its components, e.g., the ``cream'' of the class ``cake''.

Table~\ref{table: sota_imagenet1000} shows the results on the large-scale dataset ImageNet-1000 in different memory settings ($M=20k$ and $M=5k$).
We can see that our CIM-based CIL improves FOSTER consistently. It is impressive that it achieves more improvements in the more strict memory setting ($M=5k$). Specifically, it boosts the average accuracy of FOSTER by $4.5$ percentage points when $M=5k$, significantly higher than that of $M=20k$ ($1.0$).

\begin{table}
\normalsize
\begin{center}
\renewcommand\arraystretch{1.0}
\setlength{\tabcolsep}{1.5mm}{
\begin{tabular}{lcccccc}
\toprule
\multirow{2.5}{*}{\textbf{Ablation Method}} & \multicolumn{2}{c}{\textit{Food-101}} && \multicolumn{2}{c}{\textit{ImageNet-100}} \\
\cmidrule{2-3} \cmidrule{5-6}
& $N$=10 & 20 && 10 & 20 \\
\midrule
1 Baseline                             & 72.72 & 66.73 && 76.55 & 72.37 \\
2 Artifact Aug.                                   & 71.38 & 66.03 && 75.63 & 71.45 \\
\cdashline{1-6}[4pt/2pt]
3 Full Comp.                                   & 73.03 & 67.38 && 76.92 & 73.26 \\
4 Random Acti.                                    & 73.10 & 67.54 && 76.88 & 73.54 \\
5 Center Acti.                                  & 73.29 & 67.88 && 76.78 & 73.82 \\
6 Class Acti.                                     & 73.76 & 68.65 && 77.21 & 74.67 \\
\cdashline{1-6}[4pt/2pt]
7 Phase-wise $\tau$                             & 73.83 & 69.17 && 77.06 & 74.78 \\
8 Joint Train                                         & 73.44 & 69.01 && 77.34 & 74.59 \\
9 BOP (ours)                                      & 74.85 & \textbf{70.20} && \textbf{77.94} & 75.23 \\
\cdashline{1-6}[4pt/2pt]
10 LastBlock Only                                   & 74.55 & 69.87 && 77.72 & 74.86 \\
11 Fg Compressed                                     & \textbf{75.02} & 70.13 && 77.87 & \textbf{75.46} \\
\bottomrule
\end{tabular}}
\end{center}
\vspace{-5mm}
\caption{Average accuracies (\%) of different ablation methods. The experiments are conducted in the LFS setting.}
\label{table_ablation_study}
\vspace{-4mm}
\end{table}

\myparagraph{Ablation Study.}
Table~\ref{table_ablation_study} shows the ablation results. \emph{First block: baselines.}  Row 1 is for the baseline FOSTER~\cite{wang2022foster}. Row 2 shows the results of adding artifact augmentation (see Section~\ref{subsection: cam_based_compression_pipeline}). It shows directly apply this augmentation does not improve and even impair the model. 
\emph{Please note the models in below blocks all use this augmentation}.
\emph{Second block: activation methods.} 
Rows 3-6 show the results of using different activation methods to compress exemplars.
Row 3 is to downsample all pixels (i.e., no region is activated). Row 4 is to randomly select activation regions. Row 5 is to activate only the center region ($\frac{1}{4}$ of the original image), while row 6 is to use naive CAM.
Comparing them to row 1, we can see that using naively compressed exemplars can improve CIL models.
Row 4 outperforms rows 3-5, validating that it is more reliable to use the model's activation to generate compressed exemplars.
\emph{Third block: optimization methods.} 
Rows 7-9 are on top of Row 6 and are the results of applying different optimization strategies.
Row 7 is to manually select $\tau$ using a held-out set ($10\%$ of the dataset). Row 8 is to jointly train CIL and CIM models (for each input batch). Row 9 is the proposed method of using a global BOP.
\emph{Fourth block: two variants of CIM-based CIL.}
Rows 10-11 are two variants of row 9. In Row 10, only the activation layers in the last block of CIM are learnable, and previous blocks use ReLU. Compared to row 9, row 10 shows slightly worse performance. Row 11 shows the version of adding a weak downsampling ($\eta^\prime=2.0$) on discriminative regions, based on which more compressed exemplars are saved. It results in comparable performance to row 9 but increases costs.

\begin{table}
\normalsize
\begin{center}
\renewcommand\arraystretch{1}
\setlength{\tabcolsep}{0.7mm}{
\begin{tabular}{lcccccccc}
\toprule
\multirow{2.5}{*}{\textbf{Method}} & \multicolumn{3}{c}{\textit{ImageNet-100}} && \multicolumn{3}{c}{\textit{ImageNet-1000}} \\
\cmidrule{2-4} \cmidrule{6-8}
& $N$=6 & 11 & 26 && 6 & 11 & 26\\
\midrule
LUCIR \emph{\small{baseline}}                                      & 71.22 & 69.67 & 67.45 && 65.23 & 62.43 & 59.88 \\
\textit{w/} Mnemonics            & 73.30 & 72.17 & 71.50 && 66.15 & 63.12 & 63.08 \\
\textit{w/} MRDC                   & 73.62 & 72.81 & 70.44 && 67.67 & 65.60 & 62.74 \\
\textit{w/} ours                                         & \textbf{74.05} & \textbf{73.76} & \textbf{72.84} && \textbf{68.03} & \textbf{66.54} & \textbf{63.77} \\
\bottomrule
\end{tabular}}
\end{center}
\vspace{-5mm}
\caption{Comparing with Mnemonics~\cite{liu2020mnemonics} and MRDC~\cite{wang2022memory}. 
We plug each of them in baseline LUCIR~\cite{hou2019lucir} for fair comparison.
}
\label{table: comparing_with_mnemonics_and_jpeg}
\vspace{-4mm}
\end{table}
\myparagraph{Comparing with Other Compression-based Methods.} Table~\ref{table: comparing_with_mnemonics_and_jpeg} shows our results comparing to two compression-based methods: Mnemonics~\cite{liu2020mnemonics} and MRDC~\cite{wang2022memory}. We can see that our method consistently outperforms them in all settings. 
This is because our method does not sacrifice the discriminativeness of exemplars while improving the number (variance) of exemplars in a phase-adaptive manner.
However, the two related methods either keep a fixed number of exemplars in the memory~\cite{liu2020mnemonics} or use uniform image compression without considering the properties of specific classes in different incremental phases~\cite{wang2022memory}.

\begin{figure}
\centering
\includegraphics[width=0.4\textwidth]{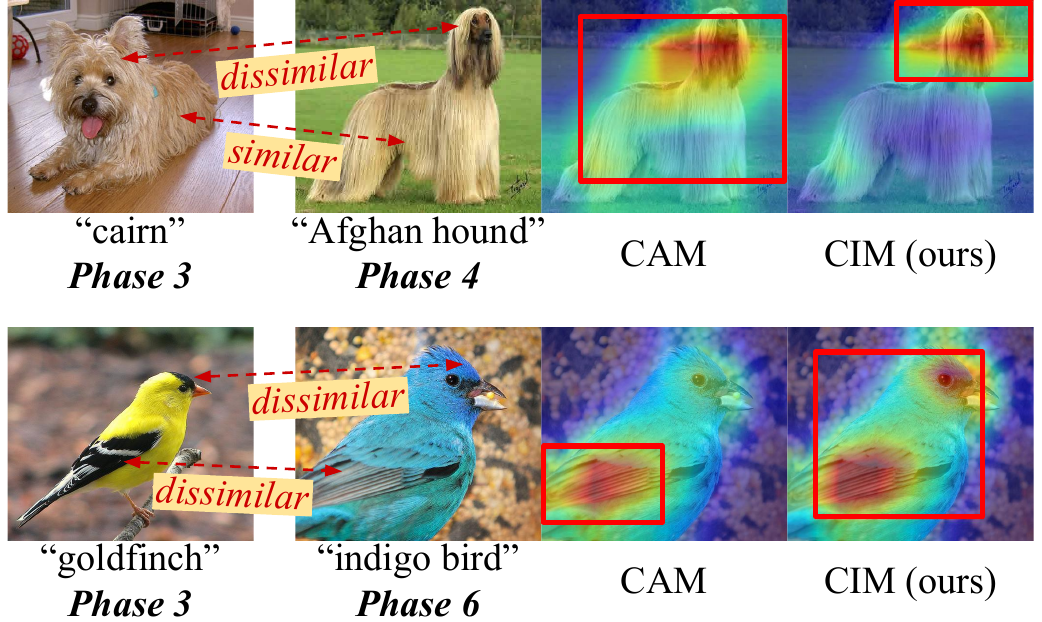}
\vspace{-2mm}
\caption{Visualizations (CAM vs. CIM). 
The experimental setting is $N$=$10$ (LFS) on ImageNet-100.
}
\label{fig_visualization}
\end{figure}

\myparagraph{Visualizations (CAM vs. CIM).} Figure~\ref{fig_visualization} gives two visualization examples, ``Afghan hound'' and ``indigo bird'', each with the activation map as well as the bounding boxes.
The first column shows their respective confusing classes appearing in earlier phases.
CIM learns to focus on the discriminative (i.e., dissimilar to confusing classes) regions.

\begin{table}
    \centering
    \vspace{-0.2cm}
    \setlength{\tabcolsep}{2.75mm}{
    \begin{tabular}{lccc}
    \toprule
    \textbf{Metric} & Small & Middle & Large \\
    \midrule
    Mean of \#Exemplars     & 39.40 & 38.30 & 34.77 \\
    \cdashline{1-4}[4pt/2pt]
    Last Acc. (\%, baseline)       & 66.13 & 68.40 & 69.93 \\
    Last Acc. (\%, ours)       & 70.00 & 71.10 & 72.26 \\
    Improvement (\%)         & +3.87 & +3.65 & +2.33 \\
    \bottomrule
    \end{tabular}
    }
    \vspace{-0.2cm}
    \caption{Results for small, middle and large objects in the setting of $N$=$10$ (LFS) on ImageNet-100.
    ``Mean of \#Exemplars'' denotes the average number of saved exemplars by our method. 
    The baseline (FOSTER~\cite{wang2022foster}) has this number as $20$ for all classes.}
    \label{table_hierarchical_compression_strategy}
    \vspace{-0.4cm}
\end{table}
\myparagraph{Results of Different-Size Objects.}
Table~\ref{table_hierarchical_compression_strategy} shows the results for small, middle, and large objects. These size categorization is according to ImageNet Object Localization Challenge~\cite{imagenetobjectlocalizationchallenge}. We calculate the bbox coverage for each class and take top $30$ classes with highest coverages as ``large'', rear $30$ classes with lowest coverages as ``small'' and the rest $40$ classes as ``middle''.
It is intriguing that our method achieves the highest improvement (over baseline) for small objects.
Our explanation is that small objects benefit more from image compression (than large ones), as their images contain more background pixels to downsample.
\section{Conclusions}
\label{section: conclusions}

We introduced a novel exemplar compression method for CIL, allowing us to save more representative exemplars but not increase memory budget. We achieved this compression by downsampling non-discriminative pixels of the image bounded by CAM masks.
To generate adaptive masks, we proposed a novel method CIM that explicitly parameterizes a mask generation model and optimizes it in an end-to-end manner across incremental phases. Our method achieves consistent performance improvements over multiple baselines and can be taken as a flexible plug-and-play module. 

\vspace{2mm}

\myparagraph{Acknowledgements.} This research was supported by the A*STAR under its AME YIRG Grant (Project No. A20E6c0101). The author gratefully acknowledges the support by the Lee Kong Chian (LKC) Fellowship fund awarded by Singapore Management University.

{\small
\bibliographystyle{ieee_fullname}
\bibliography{main}

\begin{thebibliography}{10}\itemsep=-1pt

\bibitem{abati2020conditional}
Davide Abati, Jakub Tomczak, Tijmen Blankevoort, Simone Calderara, Rita
  Cucchiara, and Babak~Ehteshami Bejnordi.
\newblock Conditional channel gated networks for task-aware continual learning.
\newblock In {\em CVPR}, 2020.

\bibitem{bochkovskiy2020yolov4}
Alexey Bochkovskiy, Chien-Yao Wang, and Hong-Yuan~Mark Liao.
\newblock Yolov4: Optimal speed and accuracy of object detection.
\newblock {\em arXiv}, 2020.

\bibitem{bossard14food101}
Lukas Bossard, Matthieu Guillaumin, and Luc Van~Gool.
\newblock Food-101 -- mining discriminative components with random forests.
\newblock In {\em ECCV}, 2014.

\bibitem{mordvintsev2014opencv}
Gary Bradski.
\newblock The opencv library.
\newblock {\em Dr. Dobb's Journal: Software Tools for the Professional
  Programmer}, 2000.

\bibitem{castleman1996digital}
Kenneth~R Castleman.
\newblock {\em Digital image processing}.
\newblock Prentice Hall Press, 1996.

\bibitem{cazenavette2022dataset}
George Cazenavette, Tongzhou Wang, Antonio Torralba, Alexei~A Efros, and
  Jun-Yan Zhu.
\newblock Dataset distillation by matching training trajectories.
\newblock In {\em CVPR}, 2022.

\bibitem{chen2022gradient}
Can Chen, Xi Chen, Chen Ma, Zixuan Liu, and Xue Liu.
\newblock Gradient-based bi-level optimization for deep learning: A survey.
\newblock {\em arXiv}, 2022.

\bibitem{chen2022recam}
Zhaozheng Chen, Tan Wang, Xiongwei Wu, Xian-Sheng Hua, Hanwang Zhang, and
  Qianru Sun.
\newblock Class re-activation maps for weakly-supervised semantic segmentation.
\newblock In {\em CVPR}, 2022.

\bibitem{cubuk2019autoaugment}
Ekin~D Cubuk, Barret Zoph, Dandelion Mane, Vijay Vasudevan, and Quoc~V Le.
\newblock Autoaugment: Learning augmentation strategies from data.
\newblock In {\em CVPR}, 2019.

\bibitem{deng2009imagenet}
Jia Deng, Wei Dong, Richard Socher, Li-Jia Li, Kai Li, and Li Fei-Fei.
\newblock Imagenet: A large-scale hierarchical image database.
\newblock In {\em CVPR}, 2009.

\bibitem{dhar2019learning}
Prithviraj Dhar, Rajat~Vikram Singh, Kuan-Chuan Peng, Ziyan Wu, and Rama
  Chellappa.
\newblock Learning without memorizing.
\newblock In {\em CVPR}, 2019.

\bibitem{dong_2020_conta}
Zhang Dong, Zhang Hanwang, Tang Jinhui, Hua Xiansheng, and Sun Qianru.
\newblock Causal intervention for weakly supervised semantic segmentation.
\newblock In {\em NeurIPS}, 2020.

\bibitem{douillard2020podnet}
Arthur Douillard, Matthieu Cord, Charles Ollion, Thomas Robert, and Eduardo
  Valle.
\newblock Podnet: Pooled outputs distillation for small-tasks incremental
  learning.
\newblock In {\em ECCV}, 2020.

\bibitem{finn2017maml}
Chelsea Finn, Pieter Abbeel, and Sergey Levine.
\newblock Model-agnostic meta-learning for fast adaptation of deep networks.
\newblock In {\em ICML}, 2017.

\bibitem{harris2020array}
Charles~R Harris, K~Jarrod Millman, St{\'e}fan~J Van Der~Walt, Ralf Gommers,
  Pauli Virtanen, David Cournapeau, Eric Wieser, Julian Taylor, Sebastian Berg,
  Nathaniel~J Smith, et~al.
\newblock Array programming with numpy.
\newblock {\em Nature}, 2020.

\bibitem{he2015prelu}
Kaiming He, Xiangyu Zhang, Shaoqing Ren, and Jian Sun.
\newblock Delving deep into rectifiers: Surpassing human-level performance on
  imagenet classification.
\newblock In {\em ICCV}, 2015.

\bibitem{he2016deep}
Kaiming He, Xiangyu Zhang, Shaoqing Ren, and Jian Sun.
\newblock Deep residual learning for image recognition.
\newblock In {\em CVPR}, 2016.

\bibitem{hou2019lucir}
Saihui Hou, Xinyu Pan, Chen~Change Loy, Zilei Wang, and Dahua Lin.
\newblock Learning a unified classifier incrementally via rebalancing.
\newblock In {\em CVPR}, 2019.

\bibitem{huang2019neural}
Shenyang Huang, Vincent Fran{\c{c}}ois-Lavet, and Guillaume Rabusseau.
\newblock Neural architecture search for class-incremental learning.
\newblock {\em arXiv}, 2019.

\bibitem{imagenetobjectlocalizationchallenge}
Imagenet object localization challenge.
\newblock
  \url{https://www.kaggle.com/competitions/imagenet-object-localization-challenge/}.

\bibitem{jain1989fundamentals}
Anil~K Jain.
\newblock {\em Fundamentals of digital image processing}.
\newblock Prentice-Hall, Inc., 1989.

\bibitem{kirkpatrick2017overcoming}
James Kirkpatrick, Razvan Pascanu, Neil Rabinowitz, Joel Veness, Guillaume
  Desjardins, Andrei~A Rusu, Kieran Milan, John Quan, Tiago Ramalho, Agnieszka
  Grabska-Barwinska, et~al.
\newblock Overcoming catastrophic forgetting in neural networks.
\newblock {\em PNAS}, 2017.

\bibitem{lee2021anti}
Jungbeom Lee, Eunji Kim, and Sungroh Yoon.
\newblock Anti-adversarially manipulated attributions for weakly and
  semi-supervised semantic segmentation.
\newblock In {\em CVPR}, 2021.

\bibitem{li2017learning}
Zhizhong Li and Derek Hoiem.
\newblock Learning without forgetting.
\newblock {\em PAMI}, 2017.

\bibitem{Liu2023Online}
Yaoyao Liu, Yingying Li, Bernt Schiele, and Qianru Sun.
\newblock Online hyperparameter optimization for class-incremental learning.
\newblock In {\em AAAI}, 2023.

\bibitem{liu2021adaptive}
Yaoyao Liu, Bernt Schiele, and Qianru Sun.
\newblock Adaptive aggregation networks for class-incremental learning.
\newblock In {\em CVPR}, 2021.

\bibitem{Liu2021RMM}
Yaoyao Liu, Bernt Schiele, and Qianru Sun.
\newblock Rmm: Reinforced memory management for class-incremental learning.
\newblock In {\em NeurIPS}, 2021.

\bibitem{liu2020mnemonics}
Yaoyao Liu, Yuting Su, An{-}An Liu, Bernt Schiele, and Qianru Sun.
\newblock Mnemonics training: Multi-class incremental learning without
  forgetting.
\newblock In {\em CVPR}, 2020.

\bibitem{loshchilov2016sgdr}
Ilya Loshchilov and Frank Hutter.
\newblock Sgdr: Stochastic gradient descent with warm restarts.
\newblock {\em arXiv}, 2016.

\bibitem{maclaurin2015gradient}
Dougal Maclaurin, David Duvenaud, and Ryan Adams.
\newblock Gradient-based hyperparameter optimization through reversible
  learning.
\newblock In {\em ICML}, 2015.

\bibitem{mccloskey1989catastrophic}
Michael McCloskey and Neal~J Cohen.
\newblock Catastrophic interference in connectionist networks: The sequential
  learning problem.
\newblock In {\em Psychology of Learning and Motivation}. Elsevier, 1989.

\bibitem{mcrae1993catastrophic}
Ken McRae and Phil~A Hetherington.
\newblock Catastrophic interference is eliminated in pretrained networks.
\newblock In {\em Proceedings of the 15h Annual Conference of the Cognitive
  Science Society}, 1993.

\bibitem{molina2019pade}
Alejandro Molina, Patrick Schramowski, and Kristian Kersting.
\newblock Pad{\'e} activation units: End-to-end learning of flexible activation
  functions in deep networks.
\newblock In {\em ICLR}, 2019.

\bibitem{nair2010rectified}
Vinod Nair and Geoffrey~E Hinton.
\newblock Rectified linear units improve restricted boltzmann machines.
\newblock In {\em ICML}, 2010.

\bibitem{parker1983comparison}
J~Anthony Parker, Robert~V Kenyon, and Donald~E Troxel.
\newblock Comparison of interpolating methods for image resampling.
\newblock {\em IEEE Transactions on Medical Imaging}, 1983.

\bibitem{paszke2019pytorch}
Adam Paszke, Sam Gross, Francisco Massa, Adam Lerer, James Bradbury, Gregory
  Chanan, Trevor Killeen, Zeming Lin, Natalia Gimelshein, Luca Antiga, et~al.
\newblock Pytorch: An imperative style, high-performance deep learning library.
\newblock In {\em NeurIPS}, 2019.

\bibitem{rebuffi2017icarl}
Sylvestre-Alvise Rebuffi, Alexander Kolesnikov, Georg Sperl, and Christoph~H
  Lampert.
\newblock {iCaRL}: Incremental classifier and representation learning.
\newblock In {\em CVPR}, 2017.

\bibitem{rusu2016progressive}
Andrei~A Rusu, Neil~C Rabinowitz, Guillaume Desjardins, Hubert Soyer, James
  Kirkpatrick, Koray Kavukcuoglu, Razvan Pascanu, and Raia Hadsell.
\newblock Progressive neural networks.
\newblock {\em arXiv}, 2016.

\bibitem{selvaraju2017grad}
Ramprasaath~R Selvaraju, Michael Cogswell, Abhishek Das, Ramakrishna Vedantam,
  Devi Parikh, and Dhruv Batra.
\newblock Grad-cam: Visual explanations from deep networks via gradient-based
  localization.
\newblock In {\em ICCV}, 2017.

\bibitem{simon2021learning}
Christian Simon, Piotr Koniusz, and Mehrtash Harandi.
\newblock On learning the geodesic path for incremental learning.
\newblock In {\em CVPR}, 2021.

\bibitem{sinha2017bilevel}
Ankur Sinha, Pekka Malo, and Kalyanmoy Deb.
\newblock A review on bilevel optimization: from classical to evolutionary
  approaches and applications.
\newblock {\em IEEE Transactions on Evolutionary Computation}, 2017.

\bibitem{sun2019meta}
Qianru Sun, Yaoyao Liu, Tat-Seng Chua, and Bernt Schiele.
\newblock Meta-transfer learning for few-shot learning.
\newblock In {\em Proceedings of the IEEE/CVF Conference on Computer Vision and
  Pattern Recognition}, pages 403--412, 2019.

\bibitem{wallace1991jpeg}
Gregory~K Wallace.
\newblock The jpeg still picture compression standard.
\newblock {\em Communications of the ACM}, 1991.

\bibitem{wang2022foster}
Fu-Yun Wang, Da-Wei Zhou, Han-Jia Ye, and De-Chuan Zhan.
\newblock Foster: Feature boosting and compression for class-incremental
  learning.
\newblock {\em arXiv}, 2022.

\bibitem{wang2022memory}
Liyuan Wang, Xingxing Zhang, Kuo Yang, Longhui Yu, Chongxuan Li, Lanqing Hong,
  Shifeng Zhang, Zhenguo Li, Yi Zhong, and Jun Zhu.
\newblock Memory replay with data compression for continual learning.
\newblock In {\em ICLR}, 2022.

\bibitem{wu2019bic}
Yue Wu, Yinpeng Chen, Lijuan Wang, Yuancheng Ye, Zicheng Liu, Yandong Guo, and
  Yun Fu.
\newblock Large scale incremental learning.
\newblock In {\em CVPR}, 2019.

\bibitem{xu2018reinforced}
Ju Xu and Zhanxing Zhu.
\newblock Reinforced continual learning.
\newblock In {\em NeurIPS}, 2018.

\bibitem{yan2021dynamically}
Shipeng Yan, Jiangwei Xie, and Xuming He.
\newblock Der: Dynamically expandable representation for class incremental
  learning.
\newblock In {\em CVPR}, 2021.

\bibitem{zenke2017continual}
Friedemann Zenke, Ben Poole, and Surya Ganguli.
\newblock Continual learning through synaptic intelligence.
\newblock In {\em ICML}, 2017.

\bibitem{zhao2020maintaining}
Bowen Zhao, Xi Xiao, Guojun Gan, Bin Zhang, and Shu-Tao Xia.
\newblock Maintaining discrimination and fairness in class incremental
  learning.
\newblock In {\em CVPR}, 2020.

\bibitem{zhou2016cam}
Bolei Zhou, Aditya Khosla, Agata Lapedriza, Aude Oliva, and Antonio Torralba.
\newblock Learning deep features for discriminative localization.
\newblock In {\em CVPR}, 2016.

\bibitem{zhou2021co}
Da-Wei Zhou, Han-Jia Ye, and De-Chuan Zhan.
\newblock Co-transport for class-incremental learning.
\newblock In {\em Proceedings of the 29th ACM International Conference on
  Multimedia}, 2021.

\end{thebibliography}
}

\clearpage
\setcounter{table}{0}
\renewcommand{\thetable}{S\arabic{table}}
\setcounter{figure}{0}
\renewcommand{\thefigure}{S\arabic{figure}}
\setcounter{section}{0}

\noindent
{\Large {\textbf{Supplementary materials}}}
\\

These supplementary materials include implementation details (\S\ref{sec_imple}), dataset details (\S\ref{sec_datas}), SOTA $95\%$ CI results (\S\ref{sec_sota}), learned PAU results (\S\ref{sec_activation_distance}), compression footprint results (\S\ref{sec_compression_footprint}), sensitivity analysis results (\S\ref{sec_sensi}), overhead analysis results (\S\ref{sec_overhead}), and hardware information (\S\ref{sec_hardw}).

\section{Implementation Details}\label{sec_imple}
\myparagraph{Downsampling Method.} \textcolor{orange}{Supplementary to Section 4.1 ``\textbf{Compression with BBox}''.}
For generating compressed exemplars, we adopted a simple downsampling method (to apply on non-discriminative pixels) called Nearest Neighbor Interpolation~\cite{parker1983comparison}. Specifically, it replaces the pixel value with that of the nearest pixel. This can be achieved by calling the \texttt{cv2.resize()} function with the \texttt{INTER\_NEAREST} flag in the standard image processing library OpenCV~\cite{mordvintsev2014opencv}.
\vspace{0.3cm}

\myparagraph{Mitigating the Effect of Compression Artifacts.} \textcolor{orange}{Supplementary to Section 4.1 ``\textbf{Compression Artifacts}''.} We mitigated the effect of compression artifacts by applying a augmentation method to new-class data $\mathcal{D}_i$ in each learning phase. 
Specifically, there were three steps. 1) For all images in $\mathcal{D}_i$, we generated the CAM-based bounding boxes at the beginning of training and updated them once per $40$ epochs.
2) In each epoch, we randomly selected a subset of $\mathcal{D}_i$. The proportion of the subset was adjusted according to the training progress---$0$ at the beginning and increased by $0.1$ every $40$ epochs.
3) Before feeding each image in the subset (into the CIL model), we downsampled (with ratio $\eta$) the pixels outside the bounding box (obtained in step 1).
\vspace{0.3cm}

\myparagraph{Compressing $\mathcal{D}_i$ into $\tilde{\mathcal{D}}_i(\phi_i)$.} \textcolor{orange}{Supplementary to Section 4.2 ``\textbf{2) Mask-level Optimization}''.} Different from the compressed pipeline introduced in Section \textcolor{red}{4.1}, compression in the inner-level optimization should involve only differentiable operations to enable gradient descent. To achieve this, we skipped the steps of adding the threshold and bounding box and obtained the compressed images by applying continuously valued masks as follows,
\begin{subequations}\label{continuous_compression}
\begin{align}
&A(\phi_i)={\omega_{i,y}^{\top}F(x;\theta_i,\phi_i)},\tag{S1a} \label{continuous_compression:a}\\
&\mathcal{M}^\textrm{CAM}(\phi_i)=\frac{A(\phi_i)-\min{(A(\phi_i))}}{\max{(A(\phi_i))} - \min{(A(\phi_i))}},\tag{S1b} \label{continuous_compression:b}\\
&\tilde{x}(\phi_i)=\mathcal{M}^\textrm{CAM}(\phi_i) \odot x + (1 - \mathcal{M}^\textrm{CAM}(\phi_i)) \odot x_\eta.\tag{S1c} \label{continuous_compression:c}
\end{align}
\end{subequations}

\myparagraph{Comparing with Other Compression-based Methods.} \textcolor{orange}{Supplementary to Table 4.} The code of plugging MRDC~\cite{wang2022memory} into PODNet~\cite{douillard2020podnet} (which shows the best results in its original paper) was not released by authors. So we rerun the results of MRDC when plugging it into LUCIR~\cite{hou2019lucir}. The experiments are conducted on the LFH setting. For a fair comparison, we apply weight transfer operations~\cite{sun2019meta} in all these experiments following Mnemonics~\cite{liu2020mnemonics}.

\section{Dataset Details}\label{sec_datas}
\textcolor{orange}{Supplementary to Section 4.1 ``\textbf{Datasets}''.} We show the details about three datasets in Table~\ref{tab_datasets}. We elaborate the image preprocessing methods applied on the three datasets in Table~\ref{tab_preprocessing}. Please note that for image preprocessing, we strictly followed~\cite{rebuffi2017icarl,hou2019lucir,douillard2020podnet,liu2020mnemonics,liu2021adaptive,yan2021dynamically,wang2022foster} for a fair comparison.
\begin{table*}[htb]
\normalsize
\vspace{-3mm}
\begin{center}
\renewcommand\arraystretch{1.2}
\setlength{\tabcolsep}{2.4mm}{
\begin{tabular}{lrrrr}
\toprule
\textbf{Dataset} & \textbf{\#Classes} & \textbf{\#Training images} & \textbf{\#Test images} & ~~~~\textbf{Avg. size} \\
\midrule
Food-101~\cite{bossard14food101}                        & 101 & 75,750 & 25,250 & 475$\times$496 \\
ImageNet-100~\cite{rebuffi2017icarl}                    & 100 & 129,395 & 5,000 & 407$\times$472 \\
ImageNet-1000~\cite{deng2009imagenet}                   & 1,000 & 1,281,167 & 50,000 & 406$\times$474 \\
\bottomrule
\end{tabular}}
\end{center}
\caption{Details of the three datasets. The ``Avg. size'' colume shows the average height$\times$width of images of each dataset.}
\label{tab_datasets}
\end{table*}
\begin{table*}[htb]
\normalsize
\begin{center}
\renewcommand\arraystretch{1.2}
\setlength{\tabcolsep}{2.4mm}{
\begin{tabular}{lll}
\toprule
\textbf{Training transformation} & \textbf{Test transformation} & \textbf{CAM transformation} \\
\midrule
\texttt{RandomResizedCrop(224),}       & \texttt{Resize(256),}        & \texttt{Resize((224,224)),} \\
\texttt{RandomHorizontalFlip(0.5),}    & \texttt{CenterCrop(224),}    & \texttt{ToTensor(),} \\
\texttt{ColorJitter(63/255),}          & \texttt{ToTensor(),}         & \texttt{Normalize().} \\
\texttt{ToTensor(),}                   & \texttt{Normalize().}        & \\
\texttt{Normalize().}                  &                            & \\
\bottomrule
\end{tabular}}
\end{center}
\caption{Uniform image preprocessing methods for the three datasets. ``CAM transformation'' denotes the preprocessing method for generating CAM~\cite{zhou2016cam}. The mean and standard deviation parameters of \texttt{Normalize()} are omitted. Note that FOSTER~\cite{wang2022foster} additionally applies AutoAugment~\cite{cubuk2019autoaugment} in the training transformation, and we followed it for fair comparison.}
\label{tab_preprocessing}
\end{table*}

\section{More Results Comparing with the SOTA}\label{sec_sota}
\begin{table*}[t]
\normalsize
\begin{center}
\renewcommand\arraystretch{1}
\setlength{\tabcolsep}{1.8mm}{
\begin{threeparttable}
\begin{tabular}{lccccccccccccccc}
\toprule
\multirow{3.5}{*}{\textbf{Method}}
& \multicolumn{7}{c}{\textit{Learning from Scratch (LFS)}} && \multicolumn{7}{c}{\textit{Learning from Half (LFH)}} \\
\cmidrule{2-8} \cmidrule{10-16}
& \multicolumn{3}{c}{\textit{Food-101}} && \multicolumn{3}{c}{\textit{ImageNet-100}} && \multicolumn{3}{c}{\textit{Food-101}} && \multicolumn{3}{c}{\textit{ImageNet-100}} \\
\cmidrule{2-4} \cmidrule{6-8} \cmidrule{10-12} \cmidrule{14-16}
& $N$=5 & 10 & 20 && 5 & 10 & 20 && 5 & 10 & 25 && 5 & 10 & 25 \\
\hline
iCaRL~\cite{rebuffi2017icarl}               & 0.58 & 1.11 & 1.50 && 0.98 & 0.64 &  -   && 1.24 & 0.96 & 0.42 && 0.86 & 1.48 & 1.37  \\
WA~\cite{zhao2020maintaining}               & 0.43 & 0.36 & 2.04 && 0.65 & 0.75 &  -   && 0.38 & 0.30 & 1.54 && 1.33 &  -   &  -    \\
PODNet~\cite{douillard2020podnet}           & 0.92 & 1.91 & 1.83 && 1.08 & 1.87 &  -   && 0.80 & 1.37 & 1.46 && 0.29 & 1.05 & 2.77   \\
AANets~\cite{liu2021adaptive}               & 0.83 & 1.56 & 0.74 && 1.31 & 1.07 & 2.01 && 1.12 & 0.48 & 1.22 && 0.53 & 0.74 & 0.81   \\
\cdashline{1-16}[4pt/2pt]
DER~\cite{yan2021dynamically}               & 0.47 & 0.62 & 0.94 && 0.46 & 0.39 &  -   && 0.56 & 0.68 &  -   && 0.51 &  -   &  -    \\
DER~\textit{w}/ ours                        & 0.28 & 0.75 & 1.10 && 0.32 & 0.24 & 0.65 && 0.17 & 0.52 &  -   && 0.41 & 0.53 &  -    \\
\cdashline{1-16}[4pt/2pt]
FOSTER~\cite{wang2022foster}                & 0.34 & 0.43 & 0.89 && 0.51 & 0.53 &  -   && 0.34 & 0.32 & 0.60 && 0.07 &  -   & 0.38  \\
FOSTER~\textit{w}/ ours                     & 0.45 & 0.22 & 1.25 && 0.36 & 0.42 & 0.54 && 0.26 & 0.17 & 0.52 && 0.44 & 0.18 & 0.61  \\
\bottomrule
\end{tabular}
\end{threeparttable}}
\end{center}
\caption{The $95\%$ confidence intervals (\%) for the results in Table~\textcolor{red}{1}.
}
\label{table: confidence_interval}
\end{table*}
\textcolor{orange}{Supplementary to Table 1.} In Table~\ref{table: confidence_interval}, we report the $95\%$ confidence intervals corresponding to the numbers in the Table~\textcolor{red}{1} of the main paper.

\section{Results of Learned PAUs}\label{sec_activation_distance}
\begin{figure}
   \center
   \includegraphics[width=0.48\textwidth]{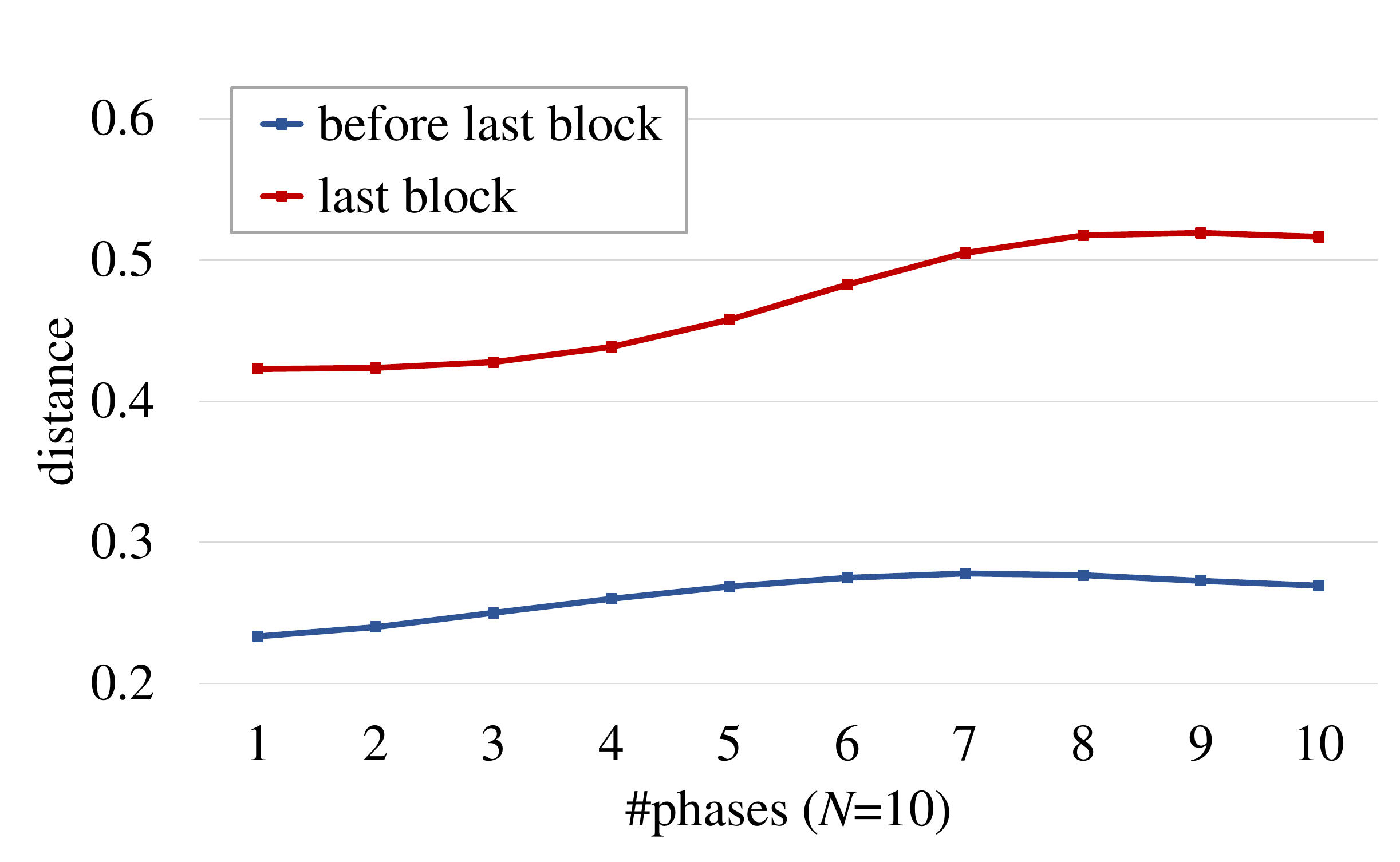}
   \caption{
   The average activation distances between ReLU~\cite{nair2010rectified} and the learned PAUs~\cite{molina2019pade} before and in the last block over all phases. The experimental setting is $N$=$10$ (LFS) on ImageNet-100. The curves are smoothed with Gaussian ($\sigma=2$).
   }
   \label{figure: activation_distance}
\end{figure} 
\textcolor{orange}{Supplementary to Section 5.2 ``Results and Analyses''.}. Figure~\ref{figure: activation_distance} shows the activation distances in different network blocks and phases. We measure the distance by $\int_{-3}^3 |f_\text{PAU}(x)-f_\text{ReLU}(x)| dx$ and use $601$ interpolated points to approximate the integration value. The learned PAUs in the last block have larger distances than those in shallow blocks. The last block mainly encodes high-level semantic information (e.g., ``body'' of ``dog''), this suggests that the learned PAUs are adjusted to focus on the most discriminative semantics. Shallow blocks learn to capture low-level features that are shareable between classification and mask generation. Therefore, the learned PAUs in these shallow blocks have little adjustment. This motivates the variant in Row 10 of Table~\textcolor{red}{3} in the main paper.

\section{Results of Compression Footprints.}\label{sec_compression_footprint}
\begin{figure}
   \center
   \includegraphics[width=0.48\textwidth]{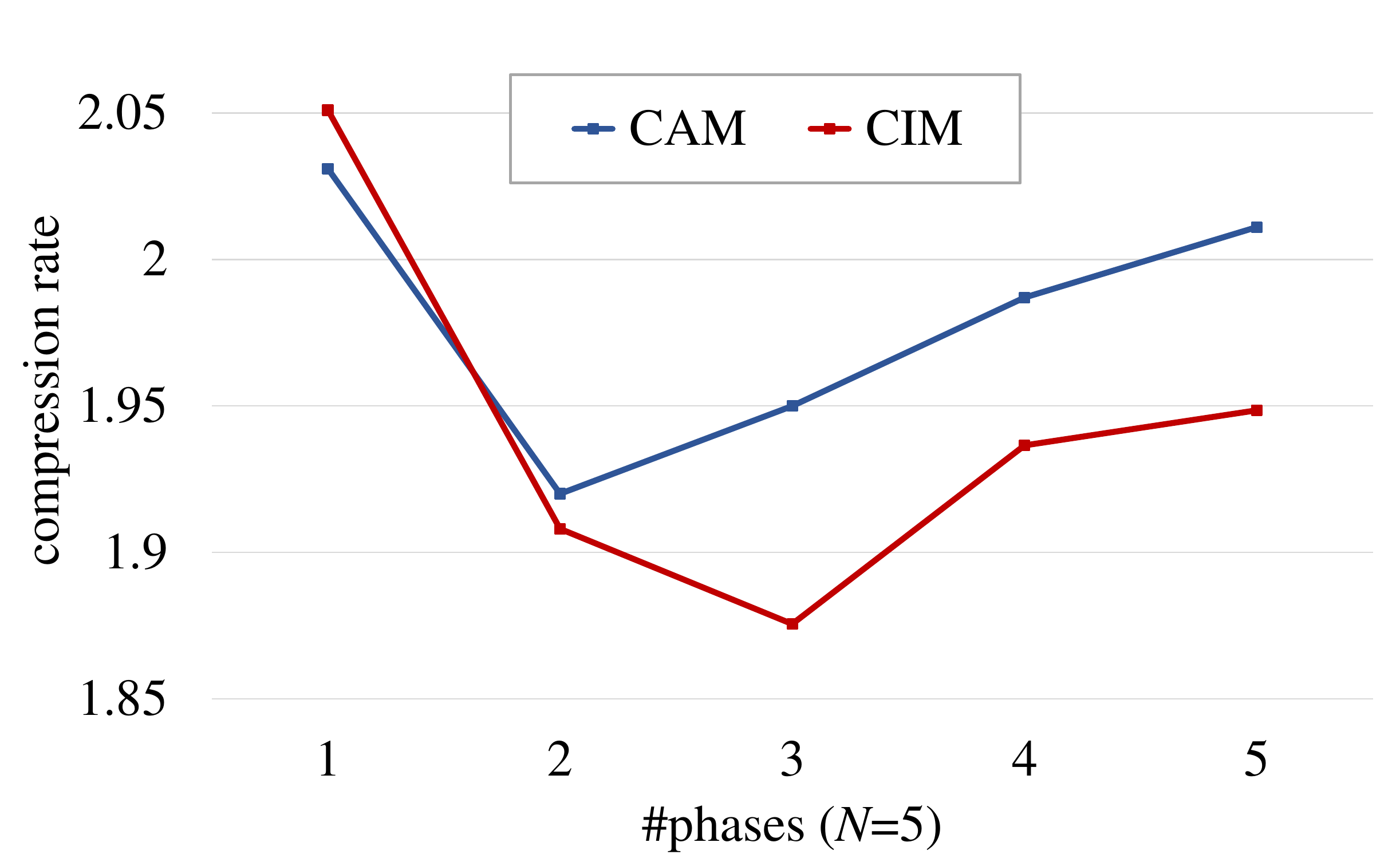}
   \caption{
   The compression rates resulted from CAM and CIM. The experimental setting is $N$=$5$ (LFS) on ImageNet-100.
   }
   \label{figure: memory_footprints}
\end{figure} 
\textcolor{orange}{Supplementary to Section 5.2 ``Results and Analyses''.}. Figure~\ref{figure: memory_footprints} provides the compression footprints along incremental phases. Comparing with CAM, CIM learns to produce more conservative compression footprints in later phases. Our explanation is that more visual cues are required to classify more classes.

\section{Results of Sensitivity Analyses}\label{sec_sensi}
\begin{figure*}[htp]
   \center
\begin{subfigure}{0.48\textwidth}
\includegraphics[width=\textwidth]{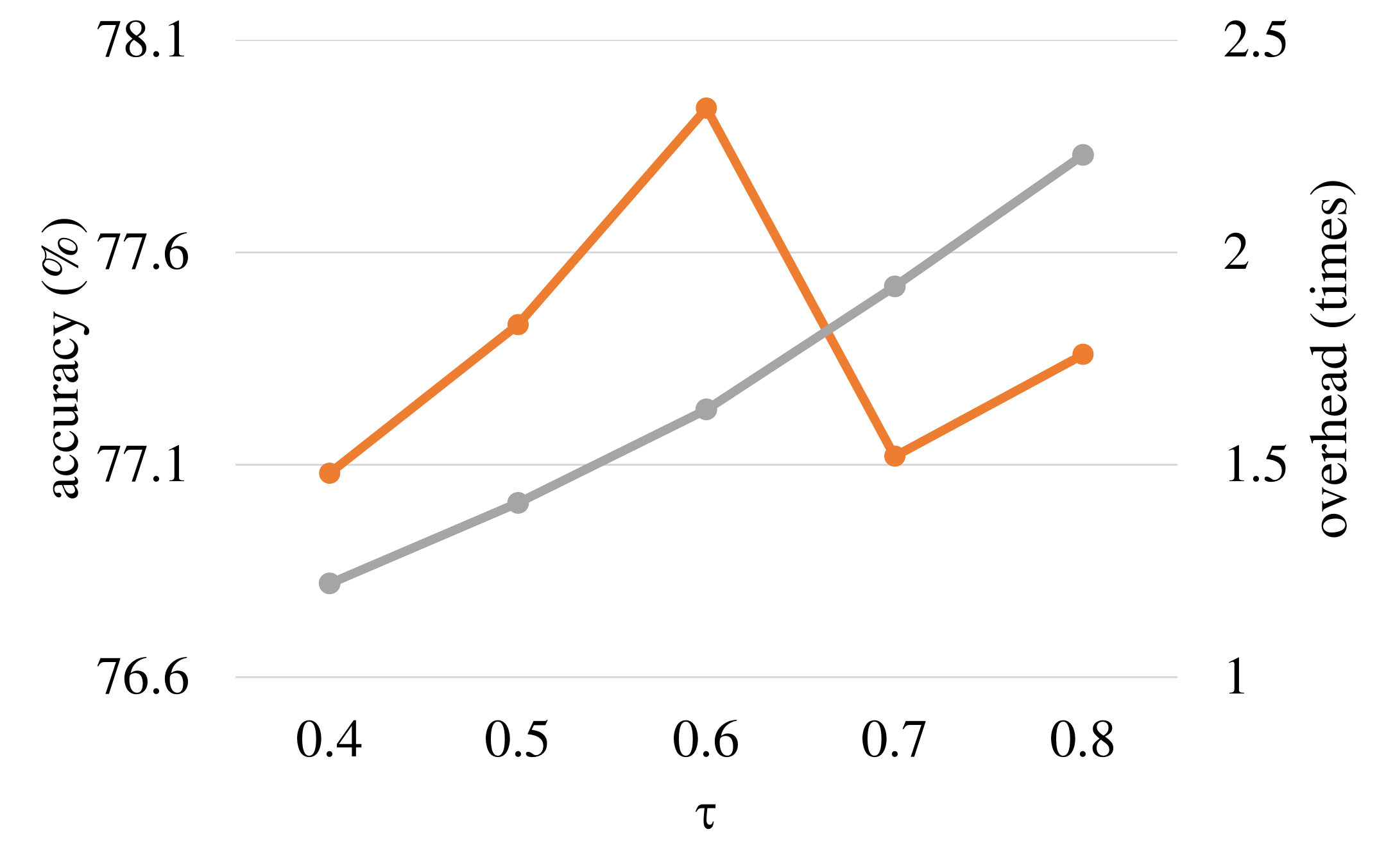}
\caption{Masking threshold $\tau$.}
\end{subfigure}
\begin{subfigure}{0.48\textwidth}
\includegraphics[width=\textwidth]{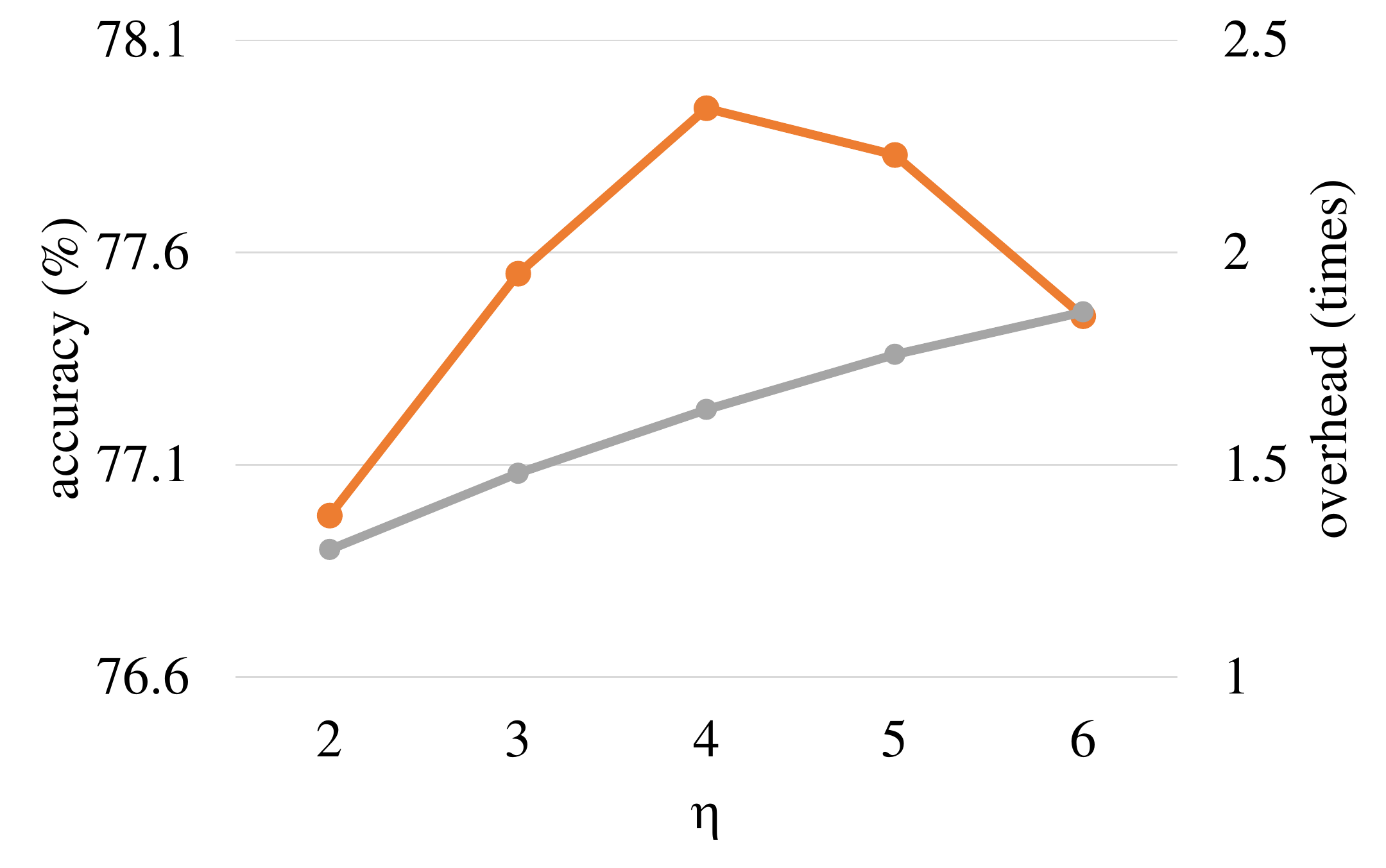}
\caption{Downsampling ratio $\eta$.}
\end{subfigure}
\begin{subfigure}{0.48\textwidth}
\includegraphics[width=\textwidth]{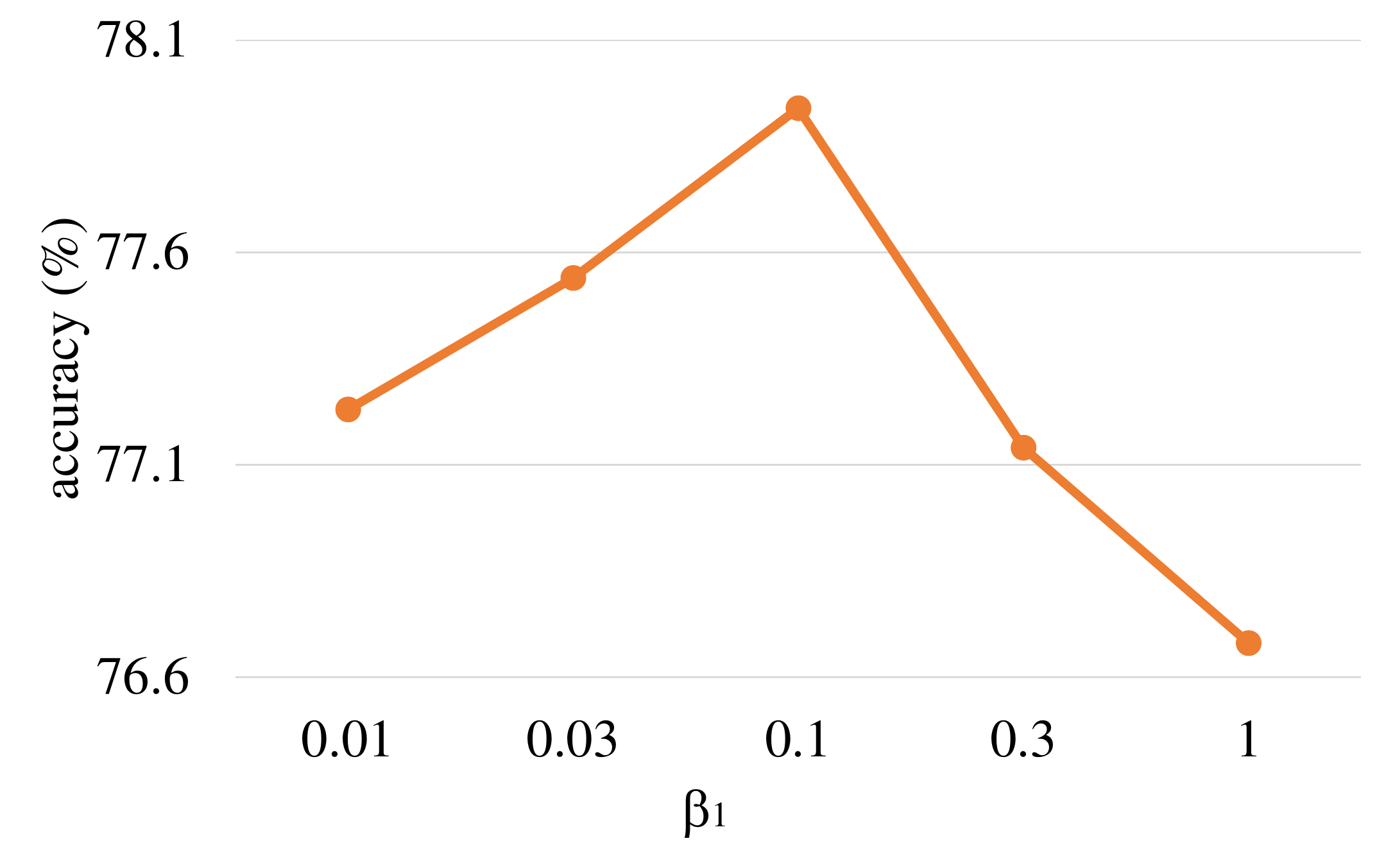}
\caption{Inner-level learning rate $\beta_1$.}
\end{subfigure}
\begin{subfigure}{0.48\textwidth}
\includegraphics[width=\textwidth]{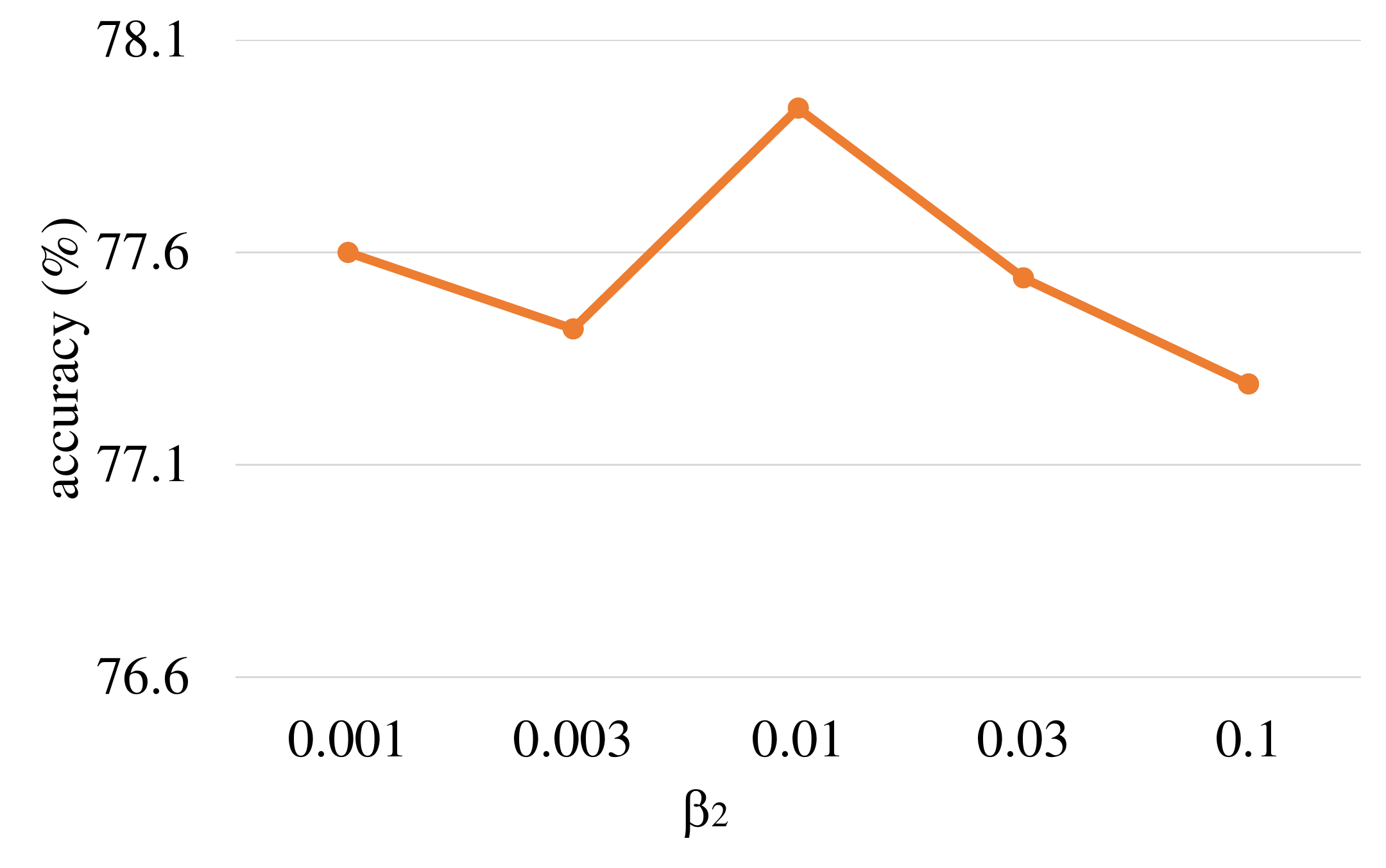}
\caption{Outer-level learning rate $\beta_2$.}
\end{subfigure}
\begin{subfigure}{0.48\textwidth}
\includegraphics[width=\textwidth]{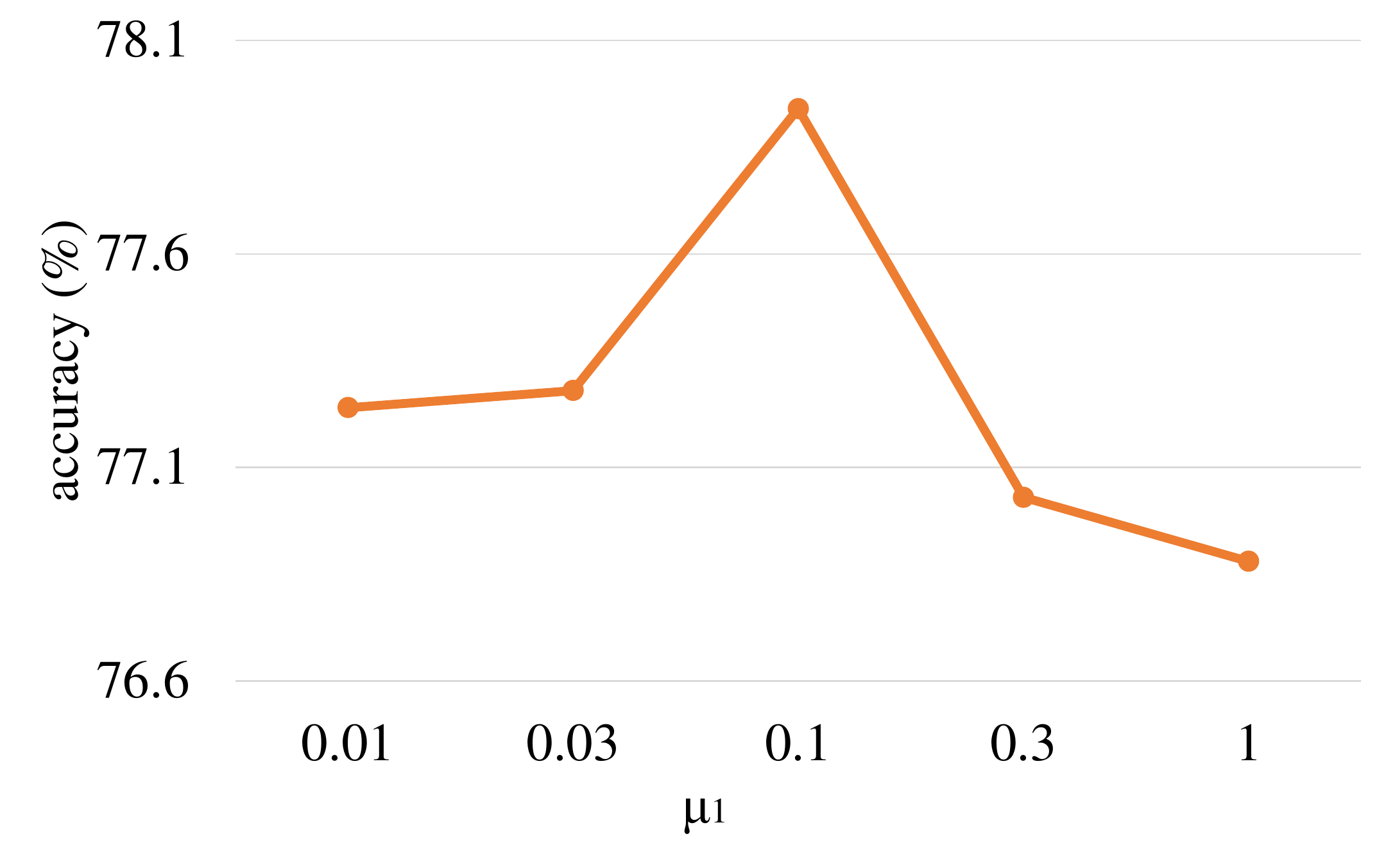}
\caption{$\ell_2$ regularization weight $\mu$.}
\end{subfigure}
\begin{subfigure}{0.48\textwidth}
\includegraphics[width=\textwidth]{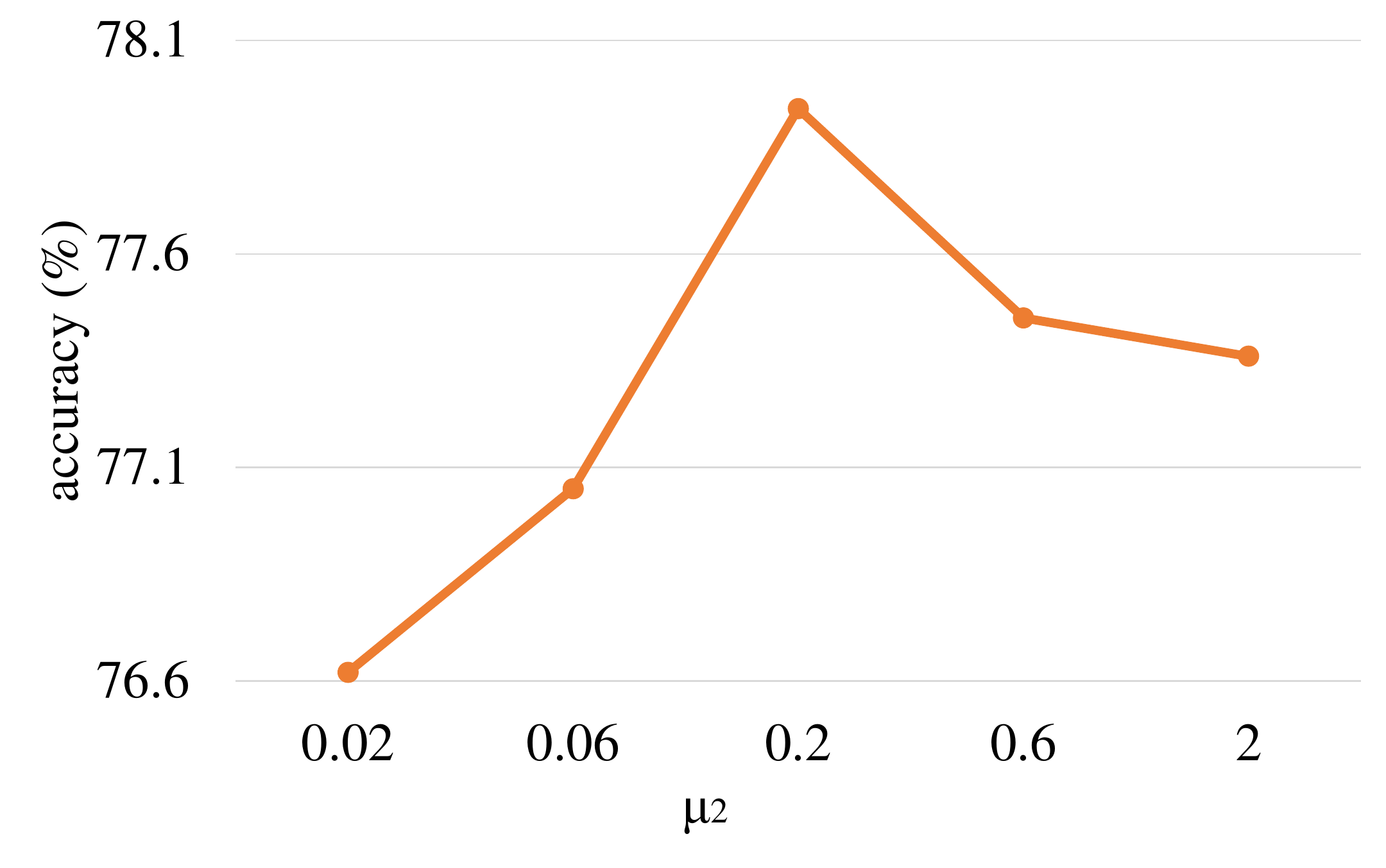}
\caption{CE regularization weight $\mu^\prime$.}
\end{subfigure}
   \caption{
   Results of hyperparameter sensitivity.
   }
   \label{figure: sensitivity}
\end{figure*} 
\textcolor{orange}{Supplementary to Section 5.1 ``Implementation Details''.} In Figure~\ref{figure: sensitivity}, we show the sensitivity analyses for CAM threshold $\tau$, downsampling ratio $\eta$, mask-level learning rates $\beta_1,\beta_2$ and two regularization weights $\mu,\mu^\prime$ on ImageNet-100~\cite{rebuffi2017icarl}. For $\tau$ and $\eta$, we also show the computation overheads.

\section{Space and Computation Overheads}\label{sec_overhead}
\textcolor{orange}{Supplementary to Section 4.2 ``Limitations''.} We elaborate on the space overhead by taking ResNet-18 as an example. We add $17$ PAUs to it, each with $10$ optimizable parameters. So we use $170$ extra parameters in total. 
This is negligible compared to $11$ million of network parameters. 
Besides, we save bbox along with exemplars in the memory. Each bbox takes around $0.01\%$ memory of a $224\times224$ RGB image.
For computation overhead, 
our method needs around $60\%$ extra computations over baseline CIL training, caused by two factors: 1)~BOP between CIL and CIM models; and 2)~training on a large number of compressed exemplars.

\section{Hardware Information}\label{sec_hardw}
-- \textbf{CPU}: AMD EPYC 7F72 24-Core Processor

-- \textbf{GPU}: $4\times$ NVIDIA GeForce RTX 3090

-- \textbf{Mem}: $8\times$ DDR4-3200 ECC RDIMM - 32GB

\end{document}